\documentclass[10pt,twocolumn,letterpaper]{article}

\usepackage{cvpr}
\usepackage{times}
\usepackage{epsfig}
\usepackage{graphicx}
\usepackage{amsmath}
\usepackage{amssymb}
\usepackage{enumerate}
\usepackage{caption}
\usepackage{subcaption}
\usepackage{comment}
\usepackage{booktabs}

\usepackage[pagebackref=true,breaklinks=true,letterpaper=true,colorlinks,bookmarks=false]{hyperref}

\cvprfinalcopy % *** Uncomment this line for the final submission

\def\cvprPaperID{****} % *** Enter the CVPR Paper ID here

\ifcvprfinal\fi
\begin{document}

%%%%%%%%% TITLE
\title{The Effects of Super-Resolution on Object Detection Performance in Satellite Imagery}

\author{Jacob Shermeyer and Adam Van Etten\\
CosmiQ Works, In-Q-Tel\\
{\tt\small \{jshermeyer,avanetten\}@iqt.org}
}

\maketitle
%\thispagestyle{empty}

%%%%%%%%% ABSTRACT
\begin{abstract}
   We explore the application of super-resolution techniques to satellite imagery, and the effects of these techniques on object detection algorithm performance. Specifically, we enhance satellite imagery beyond its native resolution, and test if we can identify various types of vehicles, planes, and boats with greater accuracy than native resolution. Using the Very Deep Super-Resolution (VDSR) framework and a custom Random Forest Super-Resolution (RFSR) framework we generate enhancement levels of $2\times$, $4\times$, and $8\times$ over five distinct resolutions ranging from 30 cm to 4.8 meters.  Using both native and super-resolved data, we then train several custom detection models using the SIMRDWN object detection framework. SIMRDWN combines a number of popular object detection algorithms (e.g. SSD, YOLO) into a unified framework that is designed to rapidly detect objects in large satellite images. This approach allows us to quantify the effects of super-resolution techniques on object detection performance across multiple classes and resolutions. We also quantify the performance of object detection as a function of native resolution and object pixel size.  
   %We find that VDSR provides an enhancement at the highest resolution, while RFSR appears more robust at lower resolutions. 
   For our test set we note that performance degrades from mean average precision (mAP) = 0.53 at 30 cm resolution, down to mAP = 0.11 at 4.8 m resolution. %We observe the greatest benefit from super-resolution for the sharpest imagery, as 
   Super-resolving native 30 cm imagery to 15 cm yields the greatest benefit; a $13 - 36\%$ improvement in mAP.  Super-resolution is less beneficial at coarser resolutions, though still provides a small improvement in performance.
   \end{abstract}

%%%%%%%%% BODY TEXT
\section{Introduction}\label{sec:intro}

The interplay between super-resolution techniques and object detection frameworks remains largely unexplored, particularly in the context of satellite or overhead imagery. Intuitively, one might assume that super-resolution methods should increase object detection performance, as an increase in resolution should add more distinguishable features that an object detection algorithm can use for discrimination. Detecting small objects such as vehicles in satellite imagery remains an exceedingly difficult task for multiple reasons \cite{van_etten_satellite_2019} and an artificial increase in resolution may help to alleviate some of these issues.  Some of the issues present include:
\begin{enumerate}

	\item Objects such as cars in satellite imagery have a small spatial extent (as low as 10 pixels) and are often densely clustered. %, occupying at most 15 pixels in the highest quality data.  	
	
	\item All objects exhibit complete rotation invariance and can have any orientation.  
	
	\item Training example frequency is low versus other disciplines.  Few datasets exist that have appropriate labels for objects within satellite imagery.  The most notable are: SpaceNet \cite{van_etten_spacenet:_2018}, %[2],  
	A Large-scale Dataset for Object DeTection in Aerial Images (DOTA) \cite{Xia_2018_CVPR}, % [3], 
	 Cars Overhead With Context (COWC) \cite{cowc}, %[4] 
	and xView \cite{lam_xview:_2018}. %[5]. 	
	
	\item Most satellite imagine sensors cover a broad area and contain hundreds of megapixels, thereby producing the equivalent of an ultra-high resolution image. For example, the native imagery used in this study was on average $\approx57$ times larger than benchmark super-resolution datasets Set5, Set14, BSD100, and Urban100.
    %{citations needed for these datasets?}.  
    When working with modern neural network architectures these images must be tiled into smaller chunks for both training and inference.
    
\end{enumerate}

Although several studies have been conducted using SR as a pre-processing step \cite{bilgazyev_sparse_2011,hennings-yeomans_simultaneous_2008,hennings-yeomans_robust_2009,shekhar_synthesis-based_2011,xu_learning-based_2017,cao_vehicle_2016,haris_task-driven_2018,dai_is_2015}, none have quantified its affect on object detection performance in satellite imagery across multiple resolutions.   This study aims to accomplish that task by training multiple custom object detection models to identify vehicles, boats, and planes in both native and super-resolved data.  We then test the models performance on the native (ground-truth) imagery and super-resolved imagery of the same Ground Sample Distance (GSD: the distance between pixels measured on the ground). Additionally, this is the first study to demonstrate the output of super-resolved 15 cm GSD satellite imagery. Although no native 15 cm satellite imagery exists for comparison, this data can be compared against coarser resolutions to test the benefits provided by super-resolution.

%For this study, super-resolution is ultimately conducted with two techniques.  The first is a convolutional neural network derived technique called Very Deep Super-Resolution (VDSR) [13].  VDSR has been featured as a baseline for the majority of recent super-resolution research and was one of the first to modify the initially proposed convolutional neural network method SRCNN[14]. The second is an approach that we have called Random-Forest Super-Resolution (RFSR) and was designed for this work; it requires minimal training time and exhibits high inference speeds.  RFSR?s most similar contemporary in recent literature is [15] and [16]. We chose to include this simpler, less computationally intensive algorithm that does not require GPUs to test its effectiveness against a near state of the art SR solution. The hypothesis is that even a simple technique may provide a benefit for object detection performance.  Ultimately, we generate enhancement levels of $2\times$, $4\times$, and $8\times$ over five distinct resolutions ranging from 30 cm to 4.8 meters.

%For object detection we chose to leverage the Satellite Imagery Multiscale Rapid Detection with Windowed Networks (SIMRDWN) pipeline[1].  This method enables the analysis of ultra-high resolution satellite imagery and the detection of small objects, many of which are only a few pixels in extent.  SIMRDWN allows an end user to run many of  the premier rapid object detection frameworks  on images of arbitrary size.  

The cost-benefit analysis of such a study is enormous. Satellite manufacturers spend the majority of their budget on the design and launch of satellites.  For example, the DigitalGlobe WorldView-4 satellite cost an estimated $\$835$ million dollars when one includes spacecraft, insurance, and launch \cite{eric_shear_ula_2016}.  Ideally, one could couple an effective SR enhancement algorithm with a smaller, cheaper satellite %with less expensive components 
that captures images in coarser resolution.  The process of capturing and subsequently enhancing coarser data could drastically reduce launch cost, expand satellite field of view,
%imaging footprint size, 
reduce the number of satellites in orbit, and improve downlink speeds between satellites and ground control stations.  

\section{Related Work}

\subsection{Super-Resolution Techniques and Application to Overhead Imagery}

Single-image Super-Resolution (SR) is the process for deriving high-resolution (HR) images from a single low-resolution (LR) image. Although super-resolution remains an ill-posed and difficult problem, recent advances in neural networks and machine learning have enabled more robust SR algorithms that exhibit effective performance. These techniques use high-resolution image pairs to learn the most likely HR features to map to the LR image features and create an output SR product.

Over the past five years, convolutional neural network approaches have been used to produce state of the art super-resolution results.  Dong et al. \cite{dong_image_2016} was the first to establish a deep learning approach with SRCNN.  This has been followed up by several successive approaches, major alterations, and improvements. Very Deep Super Resolution (VDSR) \cite{kim_accurate_2016} exhibited state of the art performance and was one of the first to modify the SRCNN approach by creating a deeper network with 20 layers to learn a residual image and transform LR images into HR images.  Developed concurrently, the Deeply-Recursive CNN (DRCN) \cite{kim_deeply-recursive_2016} introduced a recursive neural network approach to super-resolve imagery.  The Deeply Recursive Residual Network (DRRN) \cite{tai_image_2017} builds upon the VDSR and DRCN advancements using a combination of the residual layers approach and recursive learning in a compact network. 

More complex methods followed, such as the Laplacian Pyramid Super-Resolution Network (LapSRN) \cite{lai_deep_2017}.  Adversarial training has also been employed and the SR Generative Adversarial Network (SRGAN) \cite{ledig_photo-realistic_2017} produces photo-realistic $4\times$ enhanced images.  The use of wider and deeper networks has also been proposed.  The most notable being Lim et al. \cite{lim_enhanced_2017}, which proposed Enhanced Deep Residual Networks (EDSR).  
Most recently, the Deep Back Projection Network (DBPN) \cite{Haris_2018_CVPR} showed state of the art performance for an $8\times$ enhancement by connecting a series of iterative up- and down-sampling stages.  Newer block based methods such as the Information Distillation Network (IDN) \cite{Hui_2018_CVPR} was developed as a compact network that could gradually extract common features for fast the reconstruction of HR images. In another example, the Residual Dense Network (RDN) \cite{Zhang_2018_CVPR} uses residual dense blocks to produce strong performance. 
%Finally, Zero-Shot \cite{Shocher_2018_CVPR} SR is an image specific solution that was shown to be adaptable to various levels of blurring and noise.

Although new and powerful single image SR techniques continue to be developed, these techniques have been infrequently applied to overhead imagery.  One of the most notable applications of super resolution to satellite and overhead imagery remains the recent paper by Bosch et al. \cite{bosch_super-resolution_2018}.  The authors analyze several sources of satellite imagery for this research and quantify their success in terms of PSNR for an $8\times$ enhancement using a GAN. In another example, %Liu et al.
\cite{liu_single_2018} use deep neural networks for simultaneous $4\times$ super-resolution and colorization of satellite imagery. Several papers \cite{xiao_super-resolution_2018,luo_video_2017,tuna_single-frame_2018,liebel_single-image_2016,pouliot_landsat_2018} modify or leverage SRCNN \cite{dong_image_2016} and/or VDSR  \cite{kim_accurate_2016} to successfully super-resolve Jilin-1, SPOT, Pleiades, Sentinel-2, and Landsat imagery.  

Ultimately, a few specific papers are direct precursors for this work:  In the first, \cite{cao_vehicle_2016} use fine resolution aerial imagery and coarser satellite imagery with a coupled dictionary learning approach to super enhance vehicles and detect them with a simple linear Support Vector Machine model.  Their results showed that object detection performance improves when using SR as a pre-processing step versus the native coarser imagery.  Xu et al. \cite{xu_learning-based_2017} use sparse dictionary learning to generate synthetic $8\times$ and $16\times$ super-resolved imagery from Landsat and MODIS image pairs.  Their results show an increase performance for land-cover change mapping when using the super-resolved imagery. Although these approaches are similar to ours, they fail to use newer neural network based approaches, and are narrower in scope. %In two other similar studies, Yuan et al. \cite{yuan_method_2014} propose target object recognition rate as a new method for evaluating the quality of optical remote sensing imagery. They propose that the percentage of correctly identified vehicles in a satellite image can be used as a standard assessment for image quality.  The higher the recognition rate, the greater the quality an image possesses. 
Finally, \cite{haris_task-driven_2018} super-resolve imagery using DBPN \cite{Haris_2018_CVPR} and detect various objects in  traditional photography using SSD \cite{SSD}. They quantify their success in terms of mAP and also add a novel element to this work of designing a loss function to optimize SR for object detection performance.  Their results show that end-to-end training of these algorithms gave a performance boost for object detection tasks, and is a promising avenue to explore for future research.

Overall, we hypothesized that SR techniques could improve object detection performance, particularly when using satellite imagery, however no such study has been conducted. To address this question, our study investigates the relationship between object detection performance and resolution, spanning five unique GSD resolutions, with six SR outputs per resolution.   Ultimately, we investigate 35 separate resolution profiles for object detection performance.

%\vspace{-4mm}
\subsection{Object Detection Techniques}
A number of recent papers have applied advanced machine learning techniques to aerial or satellite imagery, yet have focused on a slightly different problem than the one we attempt to address. For example, \cite{rs_obj_det_slow} demonstrated the ability to localize objects in overhead imagery; yet application to larger areas would be problematic, with an inference speed of 10 - 40 seconds per $1280 \times 1280$ pixel image chip.
% this approach will not scale to large area inference. 
Efforts to localize surface to-air-missile sites \cite{samfinder} with satellite imagery and sliding window classifiers work if one only is interested in a single object size of hundreds of meters. Running a sliding window classifier across a large satellite image to search for small objects of interest quickly becomes computationally intractable, however, since multiple window sizes will be required for each object size. For perspective, one must evaluate over one million sliding window cutouts if the target is a 10 meter boat in a DigitalGlobe image. Application of rapid object detection algorithms to the remote sensing sphere is still relatively nascent, as evidenced by the lack of reference to SSD \cite{SSD}, Faster-RCNN \cite{faster_rcnn}, R-FCN \cite{rfcn}, or YOLO \cite{yolov1} in a recent survey of object detection in remote sensing \cite{rs_obj_det_survey}. While tiling a large image is still necessary, the larger field of view of these frameworks (a few hundred pixels) compared to simple classifiers (as low as 10 pixels) results in a reduction in the number of tiles required by a factor of over 1000. This reduced number of tiles yields a corresponding marked increase in inference speed. In addition, object detection frameworks often have much improved background differentiation (compared to  sliding window classifiers) since the network encodes contextual information for each object. 

As we seek to study the effect of super-resolution on object detection performance in real-world satellite imagery, %object detection studies, 
and for all of the reasons listed above - rapid object detection frameworks are the logical choice for this study.  
%Many of the best object detection frameworks can now be explored with the TensorFlow Object Detection API \cite{tf_obj_det}, and YOLO  is another popular alternative.  While the performance of all these frameworks is impressive, none can come remotely close to ingesting the hundreds of megapixel input sizes typical of satellite imagery.  %In addition, the small object size and complete rotation invariance of satellite imagery present problems.  
The premier rapid object detection algorithms (SSD, Faster-RCNN, R-FCN, and a modified version of YOLO %\cite{yolo9000} 
called YOLT \cite{yolt}) were recently incorporated into the unified framework of SIMRDWN \cite{van_etten_satellite_2019} that is optimized for ingesting satellite imagery, typically several hundred megapixels in size. The SIMRDWN paper reported the highest performance stemmed from the YOLT algorithm, followed by SSD, with Faster R-CNN and RFCN significantly behind.  %Therefore, we opt to utilize the YOLT and SSD models within SIMRDWN for this study.%Object detection also remains a highly active field of research, and four of the best object detection frameworks include Faster R-CNN, R-FCN, SSD, and YOLO.  Object detection review and perhaps a performance breakdown?  Could also add obj detection performance on satellite imagery results to section 2.4.

\section{Dataset}

\begin{figure}[b]
  \begin{subfigure}[b]{0.32\columnwidth}
    \includegraphics[width=\linewidth]{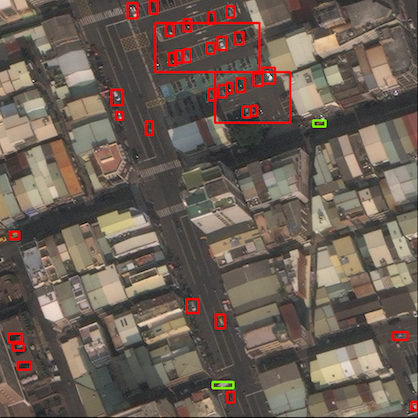}
 \caption{}
    \label{fig:1}
  \end{subfigure}
	\hfill %
  \begin{subfigure}[b]{0.32\columnwidth}
    \includegraphics[width=\linewidth]{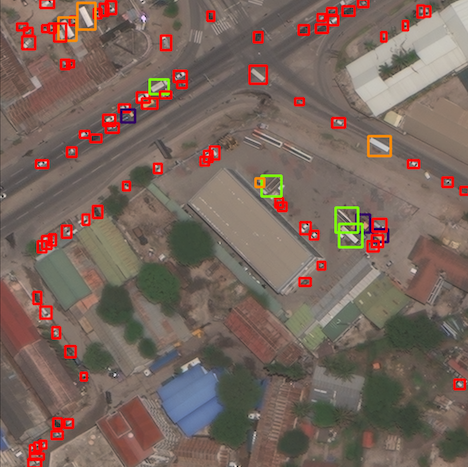}
  \caption{}
    \label{fig:2}
  \end{subfigure}
	\hfill %
  \begin{subfigure}[b]{0.32\columnwidth}
    \includegraphics[width=\linewidth]{figs/xview_crummy6_2.png}
  \caption{}
    \label{fig:3}
  \end{subfigure}
 	 \hfill %
 \vspace{-2mm}%Put here to reduce too much white space after your table 
 \caption{Issues with xView ground truth labels.  Red = car, green=truck, orange=bus, yellow=airplane, purple=boat.  Note, the incorrectly sized cars in (a), the erroneous ``boat'' ground truth labels in (b), and the missing cars in (c).}
 \label{fig:xview_goofs}
 \vspace{-4mm}%Put here to reduce too much white space after your table 
\end{figure}

\begin{comment}

\begin{figure}[htp]
  \centering
  \subfigure[random caption 1]{\includegraphics[scale=0.38]{figs/xview_crummy4.png }}\quad
  \subfigure[random caption 2]{\includegraphics[scale=0.38]{figs/xview_crummy6.png }}
\end{figure}

\begin{figure}%
\centering
\begin{subfigure}{.5\textwidth}
  \centering
  \includegraphics[width=.4\linewidth]{figs/xview_crummy4.png}
  \caption{Red = car, green = truck, orange = bus.}
  \label{fig:sub1}
\end{subfigure}%
\begin{subfigure}{.5\textwidth}
  \centering
  \includegraphics[width=.4\linewidth]{figs/xview_crummy6.png}
  \caption{Red = car, green = truck, orange = bus.}
  \label{fig:sub2}
\end{subfigure}
\caption{A figure with two subfigures}
\label{fig:test}
\end{figure}
\end{comment}

\begin{comment}

\begin{figure}
\centering
\begin{subfigure}{.5\textwidth}
  \centering
  \includegraphics[width=.4\linewidth]{image1}
  \caption{A subfigure}
  \label{fig:sub1}
\end{subfigure}%
\begin{subfigure}{.5\textwidth}
  \centering
  \includegraphics[width=.4\linewidth]{image1}
  \caption{A subfigure}
  \label{fig:sub2}
\end{subfigure}
\caption{A figure with two subfigures}
\label{fig:test}
\end{figure}

\begin{figure}
\centering
\begin{minipage}{.5\textwidth}
  \centering
  \includegraphics[width=.4\linewidth]{image1}
  \captionof{figure}{A figure}
  \label{fig:test1}
\end{minipage}%
\begin{minipage}{.5\textwidth}
  \centering
  \includegraphics[width=.4\linewidth]{image1}
  \captionof{figure}{Another figure}
  \label{fig:test2}
\end{minipage}
\end{figure}

\end{comment}

The xView Dataset \cite{lam_xview:_2018} was chosen for the application of super-resolution techniques and the quantification of object detection performance.  Imagery consists of 1,415 km$^2$ of DigitalGlobe WorldView-3 pan-sharpened RGB imagery at 30 cm native GSD resolution spread across 56 distinct global locations and 6 continents (sans Antarctica).  The labeled dataset for object detection contains 1 million object instances across 60 classes annotated with bounding boxes, including various types of buildings, vehicles, planes, trains, and boats.  
%For our purposes, we ultimately discarded larger objects such as buildings and aggregated the dataset into five specific transportation classes.  We discard classes such as ``Building,'' ``Hangar,'' and ``Vehicle Lot'' because for these objects we find that for foundational mapping efforts buildings and infrastructure projects are better represented by polygonal labels rather than bounding boxes (cite spacenet). 
For our purposes, we ultimately discarded  classes such as ``Building,'' ``Hangar,'' and ``Vehicle Lot'' because we found that such objects are better represented by polygonal labels rather than bounding boxes for foundational mapping \cite{van_etten_spacenet:_2018} purposes.

\begin{figure}%[t]
\begin{center}
\includegraphics[width=1.0\linewidth]{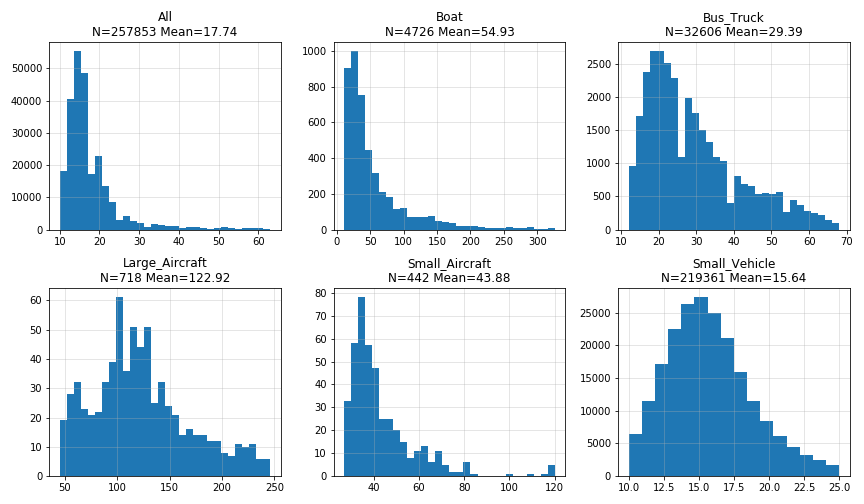}
\end{center}
\vspace{-4mm}%Put here to reduce too much white space after your table 
\caption{Object size histograms (in pixels), recall that each pixel is 30 cm in extent.}
\label{fig:sizes}
\vspace{-4mm}%Put here to reduce too much white space after your table 
\end{figure}

We chose an aggregation schema due to inconsistent labeling within the dataset.  Unfortunately, many objects are mislabeled or simply missed by labelers (see Figure \ref{fig:xview_goofs}).  This leads to an increase in false positive detection rates and objects being inaccurately tagged as mis-classifications after inference. In addition, many xView classes have a very low number of training examples (e.g. Truck w/Liquid has only 149 examples) that are poorly differentiated from similar classes (e.g. Truck w/Box has 3653 examples and looks very similar to Truck w/Liquid).  The question of how many training examples are necessary to disentangle similar classes is beyond the scope of this paper.   %, so we decided to collapse similar categories. Aggregation improves scores and makes the dataset reasonable to use for evaluation. 

\begin{table}
%\resizebox{\textwidth}{!}{%
\small%\tiny
\begin{center}
\begin{tabular}{lllll}
\hline
Category & Mean Size & \multicolumn{3}{c}{Counts} \\
%Category & Mean Size (m) & Train & Test & Total \\
& (meters) &  Train & Test & Total \\
\hline\hline
Boat & 16.5 & 2379 & 2347 & 4726 \\
Large Aircraft & 36.9 & 424 & 294 & 718 \\
Small Aircraft & 13.2 & 264 & 178 & 442 \\
Bus/Truck & 8.8 & 19337 & 13269 & 32606 \\
Small Vehicle & 4.7 & 129438 & 89923 & 219361 \\
%Large Vehicle & 19337 & 13269 & 32606 \\
\hline
\end{tabular}%}
\end{center}
\vspace{-4mm}%Put here to reduce too much white space after your table 
\caption{Object Counts}
\label{tab:xview_data}
\vspace{-4mm}%Put here to reduce too much white space after your table 
\end{table}

Our classes ultimately consist of the following (original xView classes listed in parentheses): Small Aircraft (Fixed-wing Aircraft, Small Aircraft), Large Aircraft (Cargo Plane), Small Vehicle (Passenger Vehicle, Small Car, Pickup Truck, Utility Truck), Bus/Truck (Bus, Truck, Cargo Truck, Truck w/Box, Truck w/Flatbed, Truck w/Liquid, Dump Truck, Haul Truck, Cement Mixer, Truck Tractor), and Boat (Motorboat, Sailboat, Yacht, Maritime Vessel, Tugboat, Barge, Fishing Vessel, Ferry).  See Table \ref{tab:xview_data} for dataset details and Figure \ref{fig:sizes} for object size histograms.

%===============

\subsection{Simulation of Optics and Sensors}\label{sec:data_prep}

All data were preprocessed consistently to simulate coarser resolution imagery and test the affects of our SR techniques on a range of resolutions. We intend our results to showcase what can be reasonably accomplished given coarser satellite imagery, rather than simply what is possible given the ideal settings (no blurring, bicubic decimation) under which most SR algorithms are introduced. We attempt to simulate coarser resolution satellite imagery as accurately as possible by simulating the optical point-spread function (PSF) and using a more robust decimation algorithm.  This is important because the optics of the telescope greatly impact the appearance of very small objects. The common practice of simply resizing an image by reducing its dimensions by a factor of two will simulate a different sensor containing $1/4$ the number of pixels; yet this approach ignores the different optics present in a properly designed telescope that would be coupled to such a sensor.  A properly designed sensor should have pixel size determined by the Nyquist sampling rate: half the size of the mirror resolution determined by the diffraction limit.  Given the cost and complexity of launching satellite imaging constellations to orbit, we assume that all imaging satellites will have properly designed sensors. We can use the assumption of Nyquist sampling to determine the PSF of the telescope optics, which can be approximated by a Gaussian of appropriate kernel size:

\begin{equation}
%IoU(A,B) = area(A \cap B) / area( A \cup B)
kernel = 0.5 \times GSD_{out} \, / \, GSD_{native}
\label{eqn:blur}
\end{equation}

%A base Gaussian sigma of 1 was chosen and then multiplied by the scale of degradation (Equation 1).  The more an image is degraded, the larger a Gaussian blur is initially applied.  Why did we choose this number?/citations  
For our study, data were degraded from the native 30 cm GSD using a variable Gaussian blur kernel to simulate the point-spread function of the satellite depending upon our desired output resolution (Equation \ref{eqn:blur}).  We then used inter-area decimation to reduce the dimensions of the blurred imagery to the appropriate output size (e.g. 60 cm imagery will have $1/4$ the number of pixels as 30 cm imagery over the same field of view). We repeat the above procedure to simulate resolutions of 60, 120, 240, and 480 cm.
%from our native resolution of 30 cm.  
The ground truth data and the outputs from the super-resolution algorithms were randomly split into training (60$\%$) and validation (40$\%$) categories for object detection. %Sixty-percent (60$\%$) of the data was selected for training the algorithms and forty-percent (40$\%$) for validation.  
The same images are contained in both the training and test sets regardless of resolution to maintain consistency when comparing validation scores.

\section{Super-Resolution Techniques}% and Parameterization}

For this study, super-resolution is conducted with two techniques  
%Ultimately, we generate enhancement 
for enhancement levels of $2\times$, $4\times$, $8\times$ over five distinct resolutions ranging from 30 cm to 4.8 meters.  We also create 15 cm GSD output imagery using the models trained to super-resolve imagery from 60 cm to 30 cm and 120 cm to 30 cm.

Our first method is a convolutional neural network derived technique called Very Deep Super-Resolution (VDSR) \cite{kim_accurate_2016}.  VDSR has been featured as a baseline for the majority of recent super-resolution research and was one of the first to modify the initially proposed convolutional neural network method SRCNN \cite{dong_image_2016}. This architecture was chosen due  its ease of implementation, ability to train for multiple levels of enhancement, use as a standard baseline when introducing new techniques, and favorable performance in the past. We use the standard network parameters as set in the original paper \cite{kim_accurate_2016} and train for 60 epochs.  We chose a patch size of 41 $\times$ 41 pixels and augment by rotations (4) and flipping (2) for eight unique combinations per patch.  This process is repeated for each enhancement level (2, 4, and 8$\times$), and each is fed into the same network for concurrent training.  Average training time for a 2, 4, and 8$\times$ enhancement on ~200 million pixel example is 55.9 hours on a single Titan Xp GPU.  Inference speed on a 544 $\times$ 544 pixel image is very fast $\approx0.2$ seconds on the same hardware, allowing for this method to easily scale to accommodate large satellite images.

The second method is an approach that we have called Random-Forest Super-Resolution (RFSR) and was designed for this work; it requires minimal training time and exhibits high inference speeds.  
RFSR is an adaptation of other random forest super-resolution techniques such as SRF  \cite{schulter_fast_2015} or SRRF \cite{huang_practical_2015} and can process both georeferenced satellite imagery or traditional photography. We chose to include this simpler, less computationally intensive algorithm that does not require GPUs to test its effectiveness against a near state of the art SR solution. The hypothesis is that even a simple technique may improve object detection performance.  

Our method uses a random forest regressor with a few standard parameters.  The number of estimators is set to 100, the maximum depth to 12, and the minimum samples to split an internal node equal to 200.  Finally, we use bootstrapping and out-of-bag samples to estimate the error and R$^2$ scores on randomly selected unseen data during training.  These parameters were finely tuned using empirical testing to maximize PSNR scores (see Section \ref{sec:metrics} for details on metrics) while maintaining minimal training time (4 hours or less per level of enhancement on a 64GB RAM CPU).  It should be noted that PSNR scores could be mildly improved using a deeper tree with more estimators, at the cost of training time.

Like several other SR techniques, RFSR is trained only using the luminance component from a YCbCr converted image.  HR images are degraded %using the same method for VDSR 
to create LR and HR image pairs. The degraded LR image is then shifted by one and then two pixels in each direction versus the HR image and then compressed into a 3-dimensional array. The original up-sampled LR image is then subtracted from the 3-D LR array, and from the HR image for a residual training schema.  This normalizes the LR stack and HR image pair and also removes homogeneous areas, emphasizing important edge effects.  After training and inference the interpolated LR image is then added back to the models' output image to create the super-resolved output. RFSR can only produce one level of enhancement (2, 4, or $8\times$) at a time. Average training time for all three enhancements on $\sim200$ million pixel examples is 10.8 hours on a 64GB RAM CPU.  Average inference speed on a $544\times$544 pixel image is 0.7 seconds for this same hardware (Table \ref{tab:sr_speeds}).

\begin{table}
%\resizebox{\textwidth}{!}{%
\small%\footnotesize%\small%\tiny
\begin{center}
\begin{tabular}{lll}
\hline
 & VDSR & RFSR \\
\hline\hline
 Inference Time \\ \; (per image)  & 0.16 seconds &	0.7 seconds \\
Training Time \\ \;  (for 2, 4, $8\times$)& 	55.9 hours &	10.8 hours \\
\hline
\end{tabular}%}
\end{center}
\vspace{-4mm}%Put here to reduce too much white space after your table 
\caption{Average inference time per $544\times$544 pixel image and training time for a set of 1,500 %544$\times$544 pixel 
images at native 30 cm GSD resolution. RFSR used a 64GB RAM CPU and VDSR used a NVIDIA Titan Xp GPU for inference and training.}
\vspace{-4mm}%Put here to reduce too much white space after your table 
\label{tab:sr_speeds}
\end{table}

\section{Object Detection Techniques}
%For object detection we chose to leverage the Satellite Imagery Multiscale Rapid Detection with Windowed Networks (SIMRDWN) pipeline[1].  This method enables the analysis of ultra-high resolution satellite imagery and the detection of small objects, many of which are only a few pixels in extent.  SIMRDWN allows an end user to run many of  the premier rapid object detection frameworks  on images of arbitrary size.  
As discussed in Section \ref{sec:intro}, advanced object detection frameworks have only recently been applied to large satellite imagery via the SIMRDWN framework.  In the SIMRDWN paper, the authors reported the highest performance stemmed from the YOLT algorithm, followed by SSD, with Faster R-CNN and RFCN significantly behind.  Therefore, we opt to utilize the YOLT and SSD models within SIMRDWN for this study.  	For the YOLT model we adopt the dense 22-layer network of \cite{yolt} with a momentum of 0.9, and a decay rate of 0.0005.  We use a $544\times544$ pixel training input size (corresponding to $164\times164$ meters).  Training occurs for 150 epochs.	For the SSD model we follow the TensorFlow Object Detection API implementation with the Inception V2 architecture.  We adopt a base learning rate of 0.004 and a decay rate of 0.95. We train for 30,000 iterations with a batch size of 16, and use the same $544\times544$ pixel input size as YOLT. For both YOLT and SSD we train models on the ``native'' imagery (original 30 cm data, the convolved and resized imagery described in Section \ref{sec:data_prep}), as well as on the outputs of RFSR and VDSR applied to the object detection training set.  This approach yields a multitude of models across the myriad architectures, super-resolution techniques, and resolutions (see Figure \ref{figs/sr_fig2_1.png}, thus enabling a detailed study of performance.

\begin{figure}[]
\begin{center}
\includegraphics[width=0.95\linewidth]{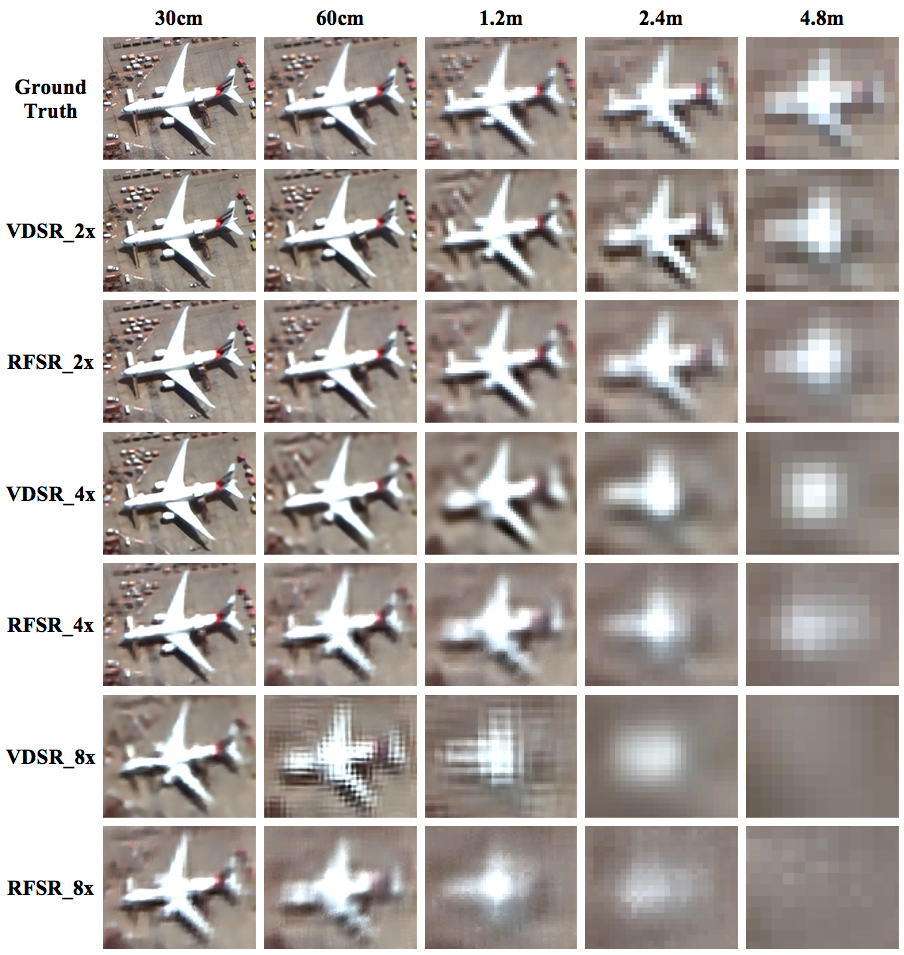}
\end{center}
\vspace{-4mm}%Put here to reduce too much white space after your table 
\caption{The effects of super-resolution on a plane and neighboring objects.  As resolution degrades super-resolution becomes a less tractable solution.}
\label{fig:sn_fig3}
\vspace{-2mm}%Put here to reduce too much white space after your table 
\end{figure}

\section{Metrics}\label{sec:metrics}
Overall, super-resolution remains an active field of research with rather limited direct focus on end application.  Typical performance metrics include Peak Signal-to-Noise Ratio (PSNR) or the Structural SIMilarity (SSIM) Index (which we report in Section \ref{sec:sr_perf}), however these measures do not quantify the enhancement to object detection performance \cite{wang_image_2004}. 
%The perceptions of humans and machines learning techniques can vary widely, and 
Although these images may be more visually appealing as a result of super-resolution, such techniques may have little impact on object detection performance. %Additionally, the range and diversity of reference image quality assessment algorithms is broad.  Kamble et al. \cite{kamble_no-reference_2015} review 74 distinct different methods, which demonstrates the multitude of ways the quality of an image can be interpreted.

For object detection metrics, we compare the ground truth bounding boxes to the predicted bounding boxes for each test image. For comparison of predictions to ground truth we define a true positive as having an intersection over union (IOU) of greater than a given threshold. An IoU of 0.5 is often used as the threshold for a correct detection, though we adopt a lower threshold of 0.25 since most of our objects are very small (e.g.: cars are only ~10 pixels in extent). This mimics Equation 5 of ImageNet \cite{imagenet}, 
%for small objects, 
which sets an IoU threshold of 0.25 for objects 10 pixels in extent. Precision-recall curves are computed by evaluating test images over a range of probability thresholds. At each of 30 evenly spaced thresholds between 0.05 and 0.95, we discard all detections below the given threshold. Non-max suppression for each object class is subsequently applied to the remaining bounding boxes; the precision and recall at that threshold is tabulated from the summed true positive, false positive, and false negatives of all test images. Finally, we compute the average precision (AP) for each object class and each model, along with the mean average precision (mAP) for each model.  One-sigma error bars are computed via bootstrap resampling, using 500 samples for each scenario.   

\section{Experimental Results}\label{src_results}

\subsection{Super-Resolution Performance}\label{sec:sr_perf}

As expected, super-resolution performance was strongest for the VDSR method, although RFSR produces  comparable results in some circumstances (Table \ref{tab:sr_perf}). As in other studies, the metrics degrade as the amount of enhancement increases.  Both techniques performed the strongest on the 60 cm imagery, likely because initial bicubic interpolation scores are high and the fact that the image resolution is situated between a coarse and fine scale where the image features are easier to detect and enhance.   

\begin{figure}[t]
\begin{center}
\includegraphics[width=0.95\linewidth]{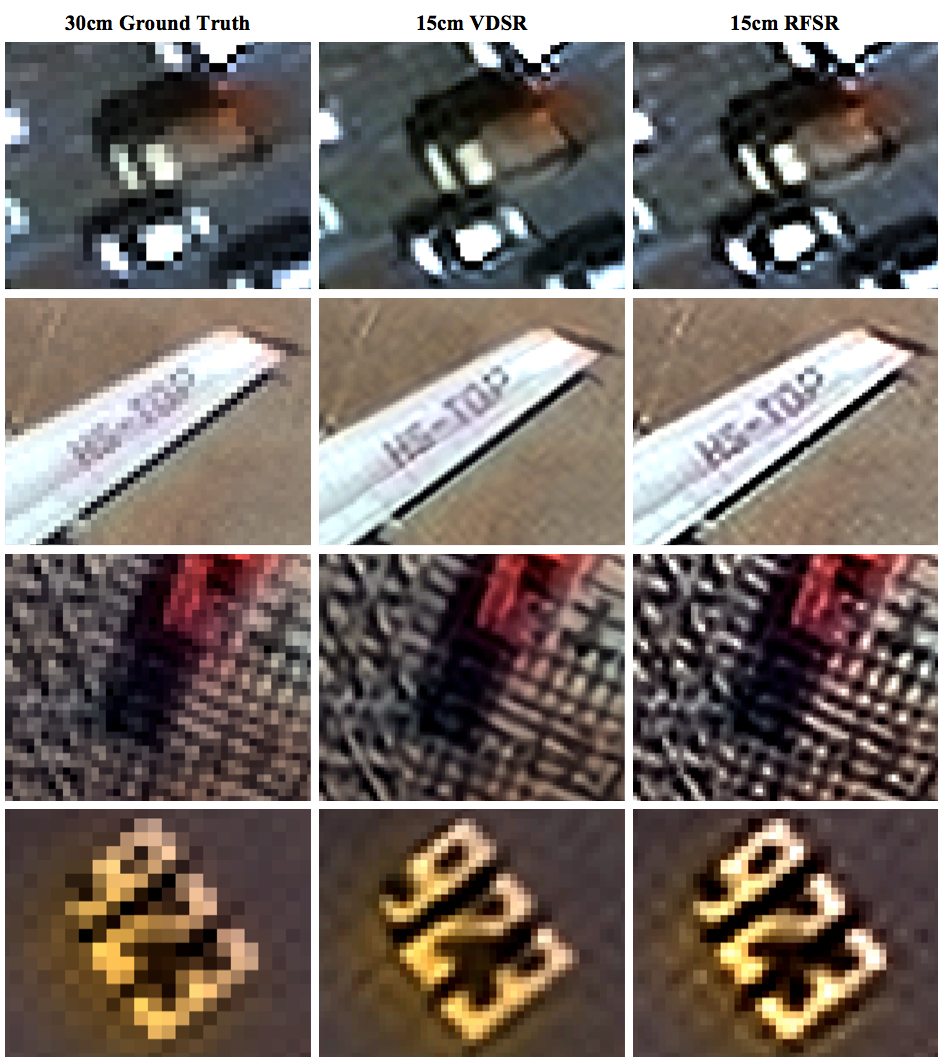}
\end{center}
\vspace{-4mm}%Put here to reduce too much white space after your table 
\caption{Examples of 15 cm GSD super-resolved output from RFSR and VDSR versus the original 30 cm GSD native imagery.}
\label{fig:sn_fig2}
\vspace{-2mm}%Put here to reduce too much white space after your table 
\end{figure}

\begin{table}
%\resizebox{\textwidth}{!}{%
\footnotesize%\small%\tiny
\begin{center}
\begin{tabular}{lllll}
\hline
GSD$_{\rm out}$ & Scale &     Bicubic &    VDSR &    RFSR \\
\hline\hline
30cm & $2\times$ & 38.68 / 0.8108 & {\bf 42.39 / 0.8925} & 39.79 / 0.8582 \\
30cm & $4\times$ & 35.86 / 0.6610 & {\bf 38.79 / 0.7795} & 35.85 / 0.7064\\
30cm & $8\times$ & 33.82 / 0.5394 & {\bf 35.69 / 0.6117} & 34.32 / 0.5874\\
\hline
60cm & $2\times$ & 41.26 / 0.9275 & {\bf 45.08 / 0.9635} &  43.03 / 0.9408\\
60cm & $4\times$ & 36.98 / 0.8082 & {\bf 40.50 / 0.8904} &  37.41 / 0.8330\\
60cm & $8\times$ & 33.99 / 0.6771 & {\bf 35.44 / 0.7293} &  33.78 / 0.6799 \\
\hline
1.2m & $2\times$ & 36.73 / 0.9151 & {\bf 39.33 / 0.9497} & 38.17 / 0.9448\\
1.2m & $4\times$ & 32.49 / 0.7738 & {\bf 35.25 / 0.8633} & 33.47 / 0.8332\\
1.2m & $8\times$ & 29.41 / 0.6097 & {\bf 30.58 / 0.6709} & 29.84 / 0.6700\\
\hline
2.4m & $2\times$ & 35.26 / 0.8848 & {\bf 41.50 / 0.9624} & 36.67 / 0.9250\\
2.4m & $4\times$ & 31.09 / 0.6898 & {\bf 33.75 / 0.8117} & 32.00 / 0.7659\\
2.4m & $8\times$ & 28.46 / 0.5004 & {\bf 30.78 / 0.6089} & 28.87 / 0.5572\\
\hline
4.8m & $2\times$ & 34.14 / 0.8404 & {\bf 37.01 / 0.9097} & 35.45 / 0.8953\\
4.8m & $4\times$ & 30.42 / 0.6079 & {\bf 33.13 / 0.7527} & 31.24 / 0.6934\\
4.8m & $8\times$ & 27.98 / 0.4013 & {\bf 30.22 / 0.5110} & 28.39 / 0.4488\\
\hline
\end{tabular}%}
\end{center}
\vspace{-4mm}%Put here to reduce too much white space after your table 
\caption{Average PSNR / SSIM scores for scale  $2\times$,  $4\times$,  and  $8\times$ across five super-resolution output GSDs. All test imagery is the xView validation dataset (281 images).  Bicubic indicates the scores if LR images are just upscaled using bicubic interpolation to match the HR image size.}
\label{tab:sr_perf}
\vspace{-4mm}%Put here to reduce too much white space after your table 
\end{table}

A few specific examples of super resolution performance are visible in Figure \ref{fig:sn_fig3}, % 1 and 2. 
%For example in Figure 1, we test the effects of our algorithm on a larger object like a plane. 
where we test the effects of our algorithm on a large object like a plane. 
Visually, VDSR and RFSR both perform strongly at 30 cm for both a $2\times$ (60 cm input $\rightarrow$ 30 cm SR output) and $4\times$ (120 cm input $\rightarrow$ 30 cm SR output) enhancement, where both the fine details of the plane, and small neighboring objects can be accurately recovered.  Recovering the plane at coarser resolutions is extremely difficult, particularly at 4.8 m with an $8\times$ enhancement.  In this case the input for the SR algorithm is 38.4 m GSD; at this resolution the satellite is simply insufficiently sensitive to resolve finer objects.  
%In figure 2, smaller objects like cars rapidly degrade and become amorphous pixelated blobs as resolution degrades. For the 30 cm data and super-resolved outputs, VDSR and RFSR perform favorably using a $2\times$ enhancement (60 cm input $\rightarrow$ 30 cm SR output). VDSR can generally recover the car shape of the 30 cm data with a $4\times$ enhancement (120 cm input $\rightarrow$ 30 cm SR output), whereas RFSR struggles to do so. 
Overall, we observe that when the imagery possesses fewer fine features to identify in coarser resolutions, algorithms are unable to hallucinate and recover all object types.
%both cars and planes.  
A different algorithm such as a GAN may be able to hallucinate visually finer features, however previous studies \cite{bosch_super-resolution_2018} have shown that these algorithms are unable to exactly recover specific features of various object types.

\begin{comment}
\begin{figure}[t]
\begin{center}
\includegraphics[width=0.9\linewidth]{figs/sr_fig1.png}
\end{center}
\vspace{-4mm}%Put here to reduce too much white space after your table 
\caption{}
\label{fig:sn_fig1}
%\vspace{-4mm}%Put here to reduce too much white space after your table 
\end{figure}
\end{comment}

Finally, in Figure \ref{fig:sn_fig2} we demonstrate the visual enhancement provided by simulated 15 cm super-resolved output from both VDSR and RFSR.  Both methods %seem comparable, 
improve the visual quality by reducing pixelization and enhancing the clarity of features and characters. RFSR appears to produce slightly brighter edge effects than VDSR.

\subsection{Object Detection Performance}

\begin{figure}[t]
\begin{center}
\includegraphics[width=0.95\linewidth]{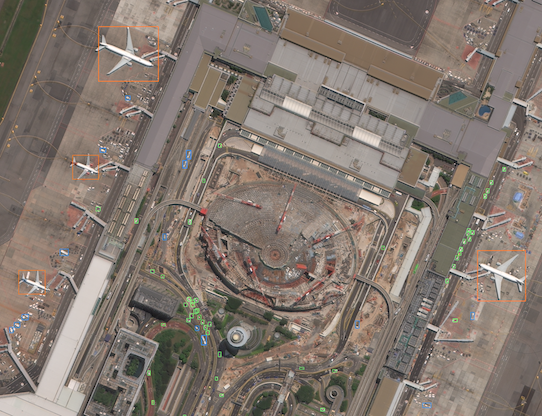}
\end{center}
\vspace{-4mm}%Put here to reduce too much white space after your table 
\caption{Example output of YOLT model at native 30 cm resolution.  Cars are in green, buses/trucks in blue, and airplanes in orange.}
\label{fig:sl0}
\vspace{-2mm}%Put here to reduce too much white space after your table 
\end{figure}

\begin{figure}
  \begin{subfigure}[b]{0.49\columnwidth}
    \includegraphics[width=\linewidth]{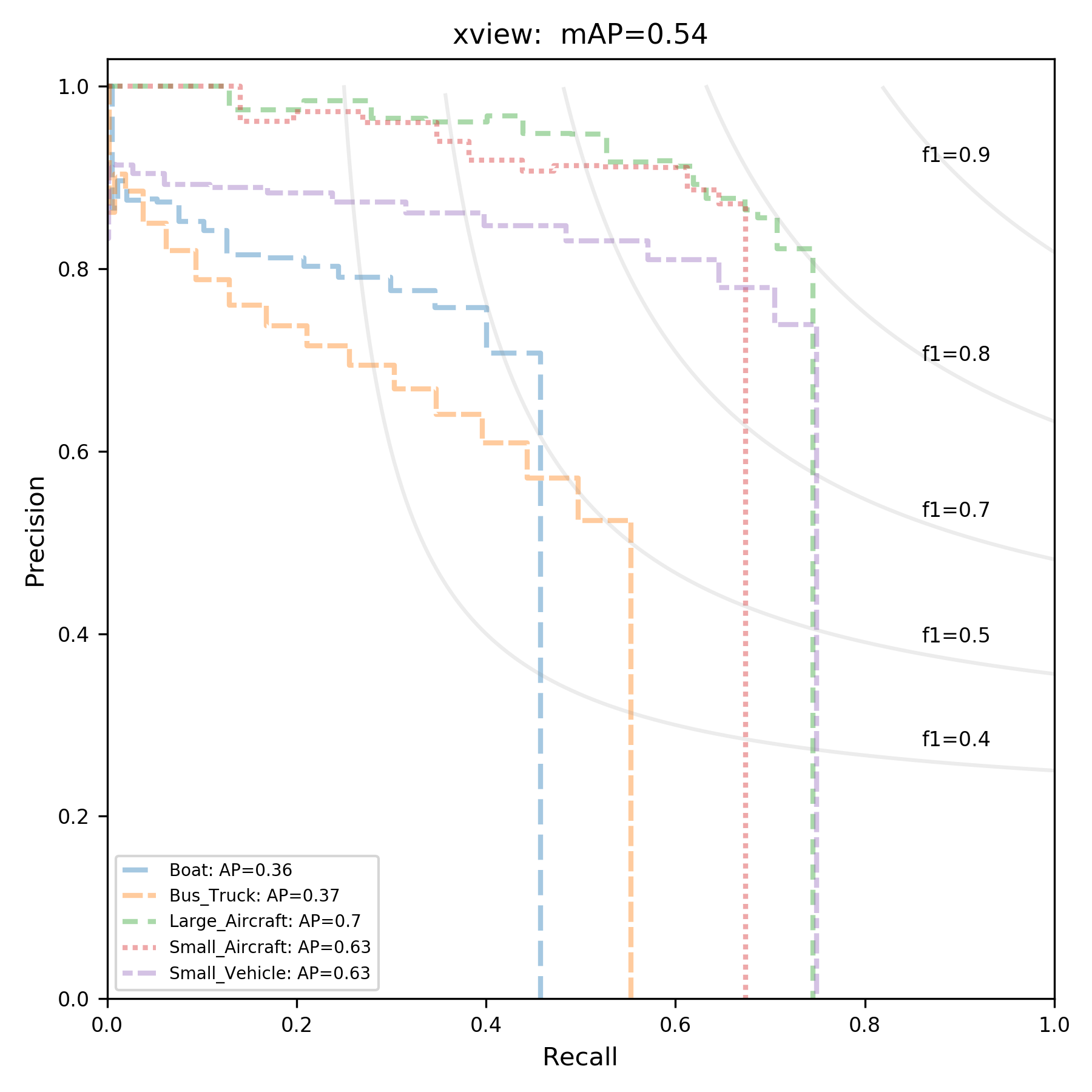}
 \caption{YOLT }
    \label{fig:1}
  \end{subfigure}
	\hfill %
  \begin{subfigure}[b]{0.49\columnwidth}
    \includegraphics[width=\linewidth]{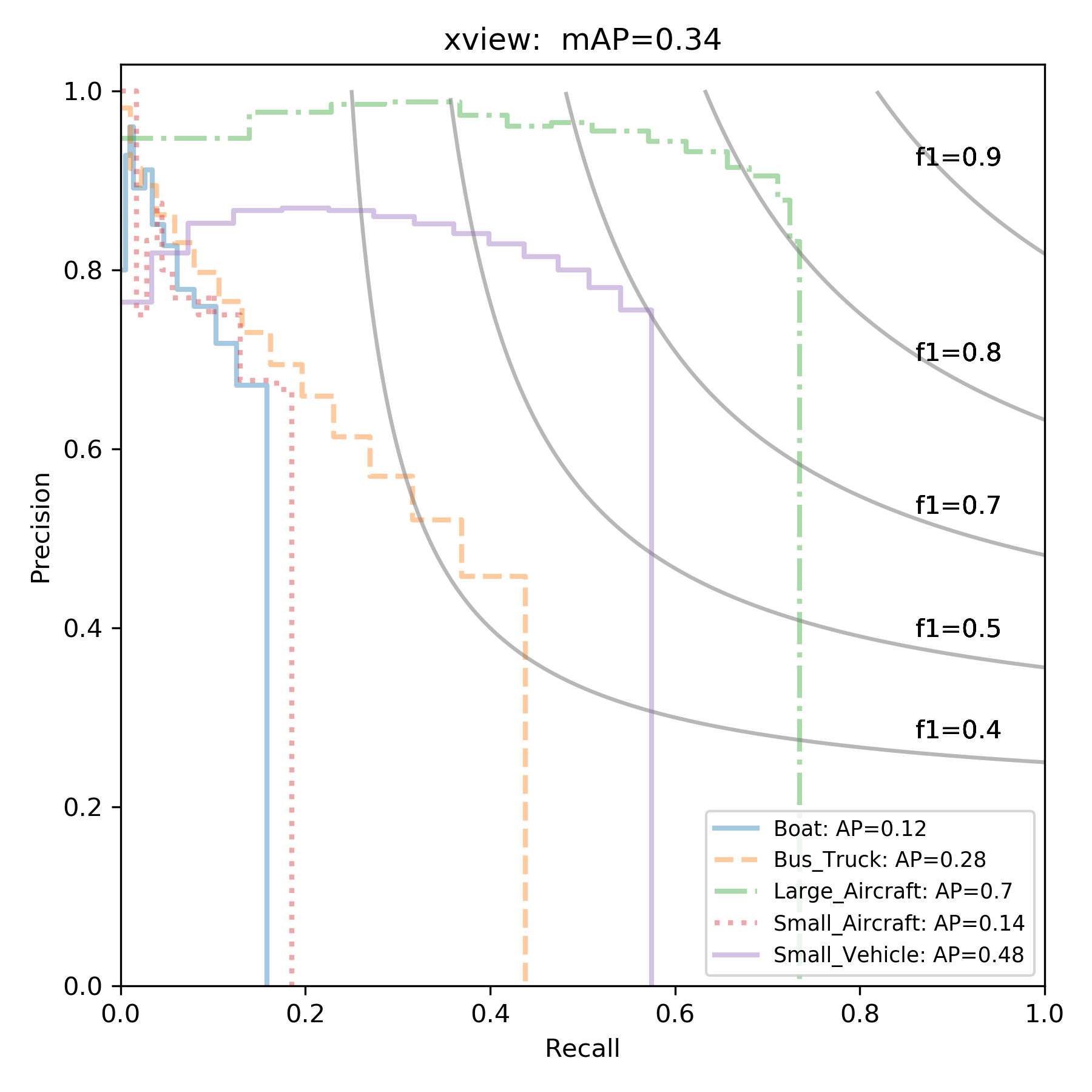}
  \caption{SSD}
    \label{fig:2}
  \end{subfigure}
	\hfill %
	\vspace{-2mm}%Put here to reduce too much white space after your table 
  \caption{Precision-recall curves for native 30 cm imagery for both YOLT and SSD.}
\label{fig:pr0}
\vspace{-4mm}%Put here to reduce too much white space after your table 
\end{figure}

For each model we compute mean average precision (mAP) performance on a 338-image test set spanning 6 continents (~632 sq. km)  at each resolution.  %We also train and test a model on native resolution imagery at double the sampling rate (with bicubic upsampling), giving a window size of 82 meters (versus 164 meters for 30 - 480 cm resolution).  We perform this test in order to disentangle the issues of resolution and window size for the 15 cm super-resolved data.  Comparing the 15 cm super-resolved predictions to the $2\times$ oversampled imagery will indicate whether any difference in performance is due to the smaller window size or the super-resolution technique at 15 cm.  
Example precision recall curves are shown in Figure \ref{fig:pr0}.  The YOLT model is clearly superior to SSD, particularly for small objects.  

\begin{figure}[t]
\vspace{-0mm}%Put here to reduce too much white space after your table 
\begin{center}
\includegraphics[width=0.999\linewidth]{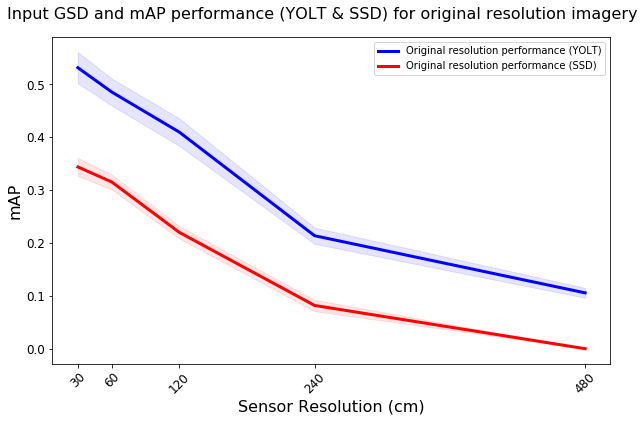}
\end{center}
\vspace{-6mm}%Put here to reduce too much white space after your table 
\caption{Performance of YOLT and SSD at the native sensor resolution for all object classes.}
\label{fig:yolt_ssd_curve}
%\vspace{-4mm}%Put here to reduce too much white space after your table 
\end{figure}

Repeating the computation shown in Figure \ref{fig:pr0} for all models allows us to determine the degradation of performance as a function of resolution, as shown in Figure \ref{fig:yolt_ssd_curve}. In this plot we display $1\sigma$ bootstrap error bars for each model group.  Results for SSD models are significantly worse than YOLT models, with a mAP of 0.30 at native 30 cm resolution.  The YOLT model (mAP = 0.53) at this resolution is 77\% better than SSD, which aligns fairly well with the findings of \cite{van_etten_satellite_2019}. Ultimately, object detection performance decreases by $22 - 27\%$ when resolution degrades from 30 cm to 120 cm, and another $73 - 100\%$ from 120 cm to 480 cm when looking across broad object classes.

%Insert table of MAP scores
\begin{table*}[t]
%\resizebox{\textwidth}{!}{%
\scriptsize%\footnotesize%\small%\tiny
\begin{center}
\begin{tabular}{lllllllll}
\hline
Model & Data           & 30 cm                          & 60 cm                          & 120 cm                         & 240 cm                         & 480 cm                         \\ \hline\hline
YOLT  & Native         & $0.53 \pm 0.03$                & $0.49 \pm 0.03$                & $0.41 \pm 0.03$                & $0.21 \pm 0.02$                & $0.11 \pm 0.01$                \\
YOLT  & RFSR $2\times$ & $0.60 \pm 0.03$ (+1.7$\sigma$) & $0.52 \pm 0.03$ (+0.7$\sigma$) & $0.39 \pm 0.03$ (-0.5$\sigma$) & $0.24 \pm 0.02$ (+1.1$\sigma$) & $0.12 \pm 0.01$ (+0.7$\sigma$) \\
YOLT  & VDSR $2\times$ & $0.60 \pm 0.03$ (+1.7$\sigma$) & $0.52 \pm 0.03$ (+0.7$\sigma$) & $0.41 \pm 0.03$ (+0.0$\sigma$) & $0.22 \pm 0.01$ (+0.4$\sigma$) & $0.13 \pm 0.01$ (+1.4$\sigma$) \\
YOLT  & RFSR $4\times$ &                                & $0.56 \pm 0.03$ (+1.6$\sigma$) & $0.40 \pm 0.03$ (-0.2$\sigma$) & $0.23\pm0.01$ (+0.9$\sigma$)   & $0.12\pm0.01$ (+0.7$\sigma$)   \\
YOLT  & VDSR $4\times$ &                                & $0.59 \pm 0.02$ (+2.8$\sigma$) & $0.39 \pm 0.03$ (-0.5$\sigma$) & $0.25\pm0.02$ (+1.4$\sigma$)   & $0.10\pm0.01$ (-0.7$\sigma$)   \\ \hline
SSD   & Native         & $0.30\pm0.01$                  & $0.32\pm0.01$                  & $0.22\pm0.01$                  & $0.08\pm0.01$                  & $0.00 \pm 0.00$                \\
SSD   & RFSR $2\times$ & $0.36\pm0.01$ (+4.2$\sigma$)   & $0.33\pm0.02$ (+0.4$\sigma$)   & $0.24\pm0.01$ (+1.4$\sigma$)   & $0.13\pm0.01$ (+3.5$\sigma$)   & $0.07\pm0.01$ (+7.0$\sigma$)   \\
SSD   & VDSR $2\times$ & $0.41\pm0.03$ (+3.5$\sigma$)   & $0.32\pm0.02$ (+0.0$\sigma$)   & $0.26\pm0.01$ (+2.8$\sigma$)   & $0.14\pm0.01$ (+4.2$\sigma$)   & $0.08\pm0.01$ (+8.0$\sigma$)   \\ \hline
\end{tabular}%}
\end{center}
\vspace{-4mm}%Put here to reduce too much white space after your table 
\caption{Performance for each data type in mAP.  For both RFSR and VDSR at each resolution we note the error and statistical difference from the baseline model (e.g. +0.5$\sigma$). The native sensor resolution of our original imagery and the input into the super-resolution models  is shown on the X-axis.  We then compare the super-resolved outputs vs. the original native imagery to test the change in object detection performance. }
\label{tab:det_perf}
\vspace{-4mm}%Put here to reduce too much white space after your table 
\end{table*}

\begin{figure}[t]
\vspace{-0mm}%Put here to reduce too much white space after your table 
\begin{center}
\includegraphics[width=0.999\linewidth]{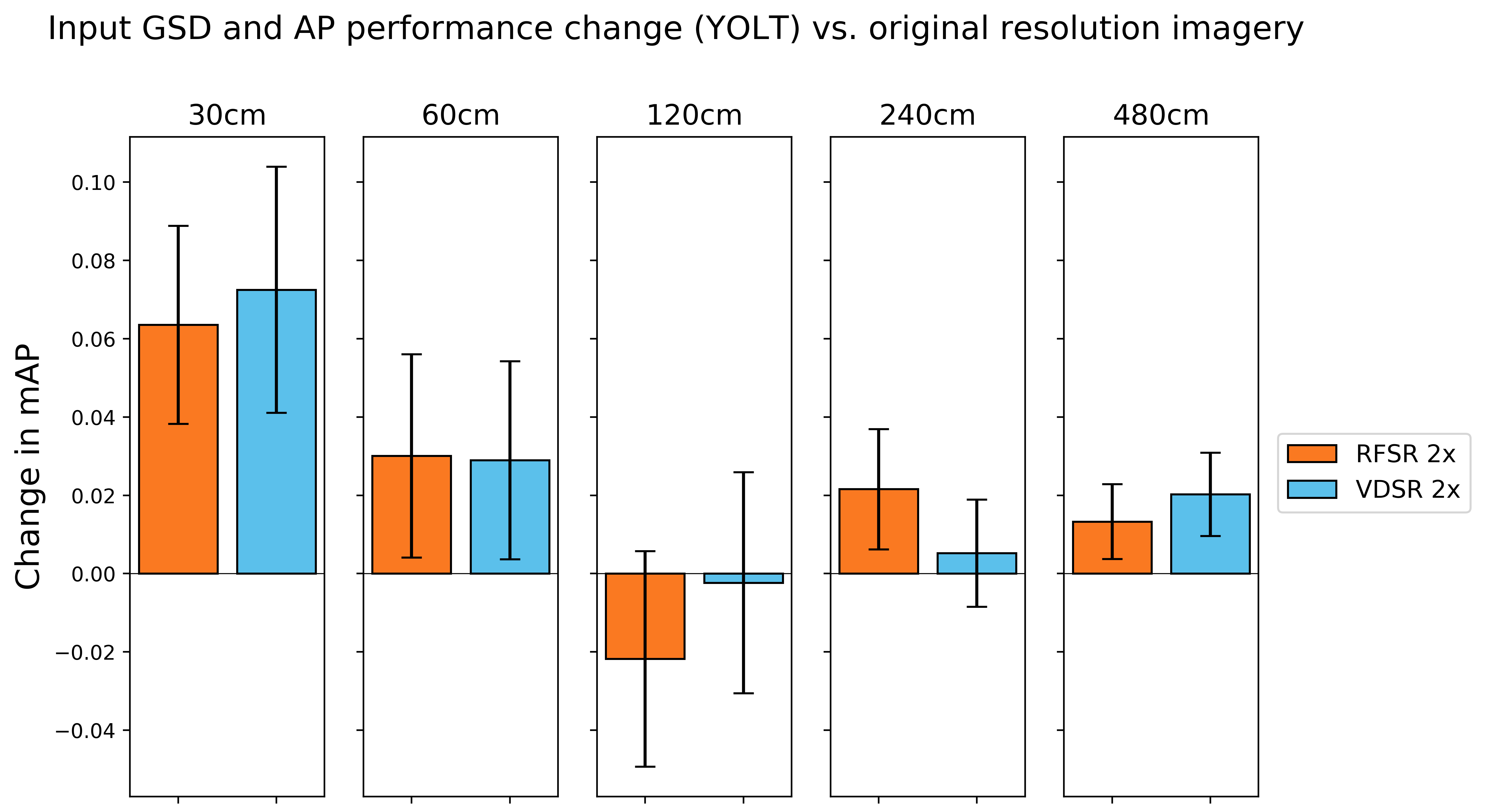}
\end{center}
\vspace{-6mm}%Put here to reduce too much white space after your table 
\caption{Performance change over original resolution (Figure \ref{fig:yolt_ssd_curve}-Blue Line) using YOLT and $2\times$ super-resolved data.}
\label{fig:yolt_bars}
\vspace{-4mm}%Put here to reduce too much white space after your table 
\end{figure}

\begin{figure}[t]
\vspace{-0mm}%Put here to reduce too much white space after your table 
\begin{center}
\includegraphics[width=0.999\linewidth]{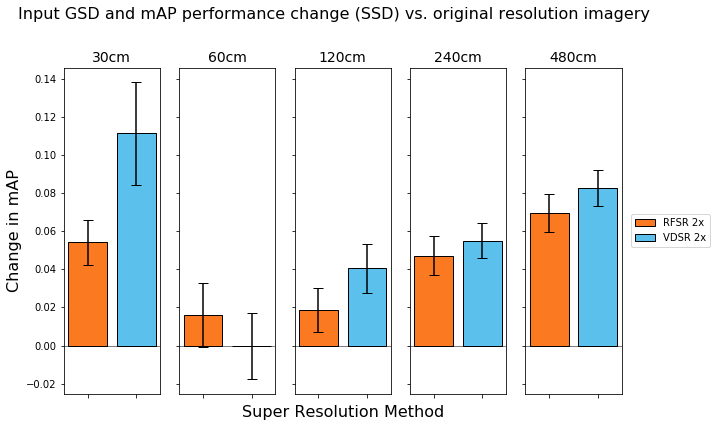}
\end{center}
\vspace{-6mm}%Put here to reduce too much white space after your table 
\caption{Performance change over original resolution (Figure \ref{fig:yolt_ssd_curve}-Red Line) using SSD and $2\times$ super-resolved data.}
\label{fig:ssd_bars}
\vspace{-4mm}%Put here to reduce too much white space after your table 
\end{figure}

\begin{figure}
\vspace{-0mm}%Put here to reduce too much white space after your table 
\begin{center}
\includegraphics[width=0.999\linewidth]{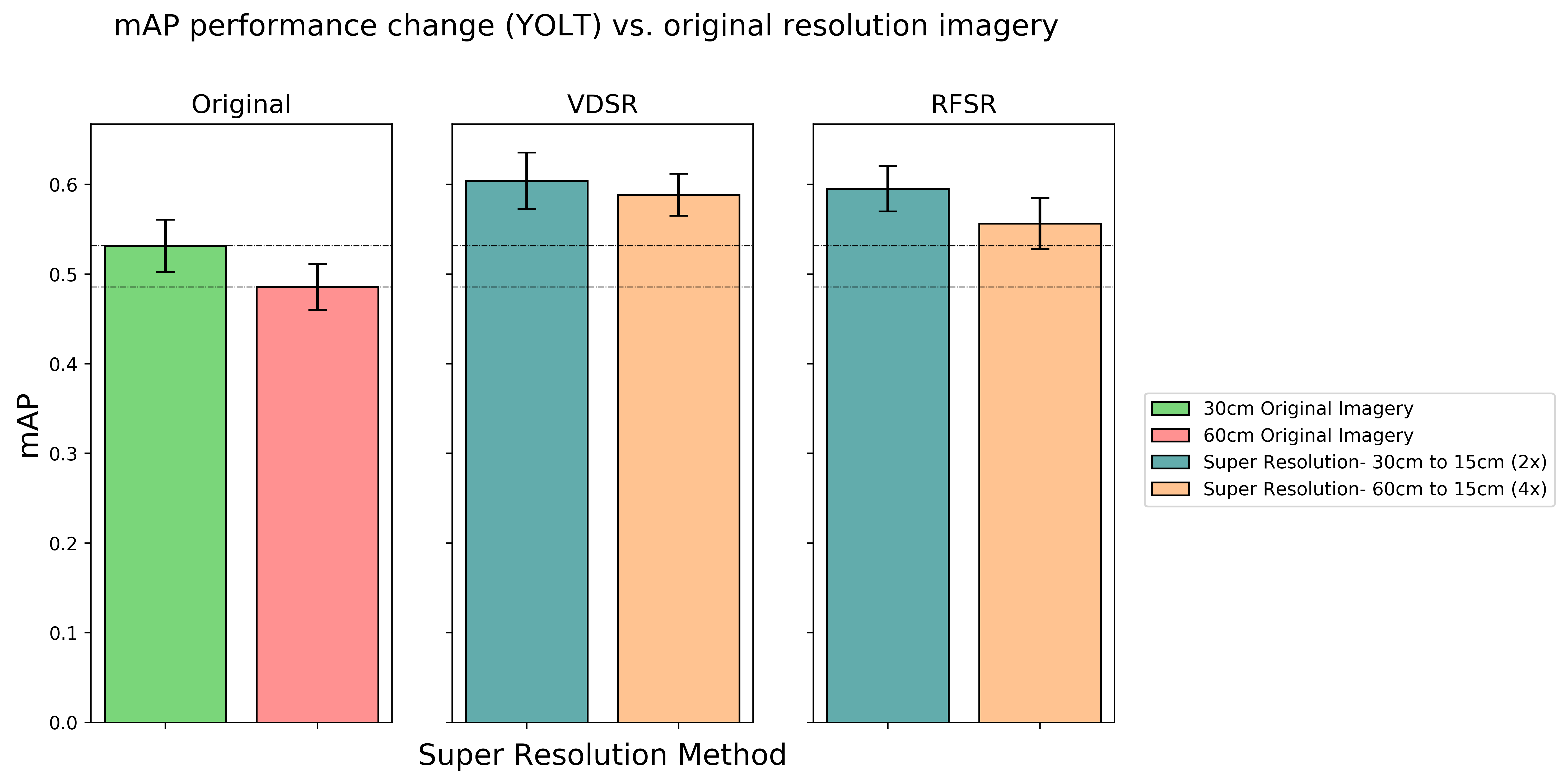}
\end{center}
\vspace{-6mm}%Put here to reduce too much white space after your table 
\caption{Performance boost of enhancing 30 and 60 cm imagery to 15 cm GSD.}
\label{fig:15cm_Boost}
\vspace{-4mm}%Put here to reduce too much white space after your table 
\end{figure}

We also plot the results of the effects of $2\times$ super-resolution models when using both YOLT and SSD  (Figures \ref{fig:yolt_bars} and \ref{fig:ssd_bars}).  When using YOLT, performance improvements are statistically significant only in the finest resolutions (Table \ref{tab:det_perf}) with comparable results for both VDSR and RFSR.  In Figure \ref{fig:15cm_Boost} we show the change in mAP versus the original 30 cm and 60 cm imagery. The largest performance boosts can be seen when enhancing imagery from 30 cm to 15 cm ($+13\%$ vs 30 cm) and 60 cm to 15 cm 
%($+14\% \, \rm{to} +20\%$ vs 60 cm).  
($14 - 20\%$ improvement vs 60 cm).  
Interestingly, enhancing imagery from 60 cm to 30 cm was much less effective than enhancing imagery from 60 to 15 cm. These findings showcase the value of super-resolution as a pre-processing step in these GSDs.  Combined with a state of the art object detection framework, super-resolution has the ability to improve detection rates beyond what is possible with the best commercially available satellite imagery.

%For YOLT, the greatest benefit is achieved at the highest resolutions, as super-resolving native 30 cm imagery to 15 cm with VDSR yields a 2.7$\sigma$ ($+20\%$) improvement.  VDSR $2\times$ yields little improvement at lower resolution, averaging a $3\%$ improvement for 60 - 480 cm, and harming performance when used to enhance imagery from 240 cm to 120 cm.  Contrastingly, RFSR provides a small but statistically insignificant boost to performance at any resolution, except when used to enhance imagery from 240 cm to 120 cm. We also train YOLT models for $4\times$ enhancement, and include results in Table \ref{tab:det_perf} for $60\rightarrow15$, $120\rightarrow30$, $240\rightarrow60$, and $480\rightarrow120$ cm.  We note similar results to the $2\times$ enhancement; further information can be found in the supplemental.

\begin{figure}
  \begin{subfigure}[b]{0.48\columnwidth}
    \includegraphics[width=\linewidth]{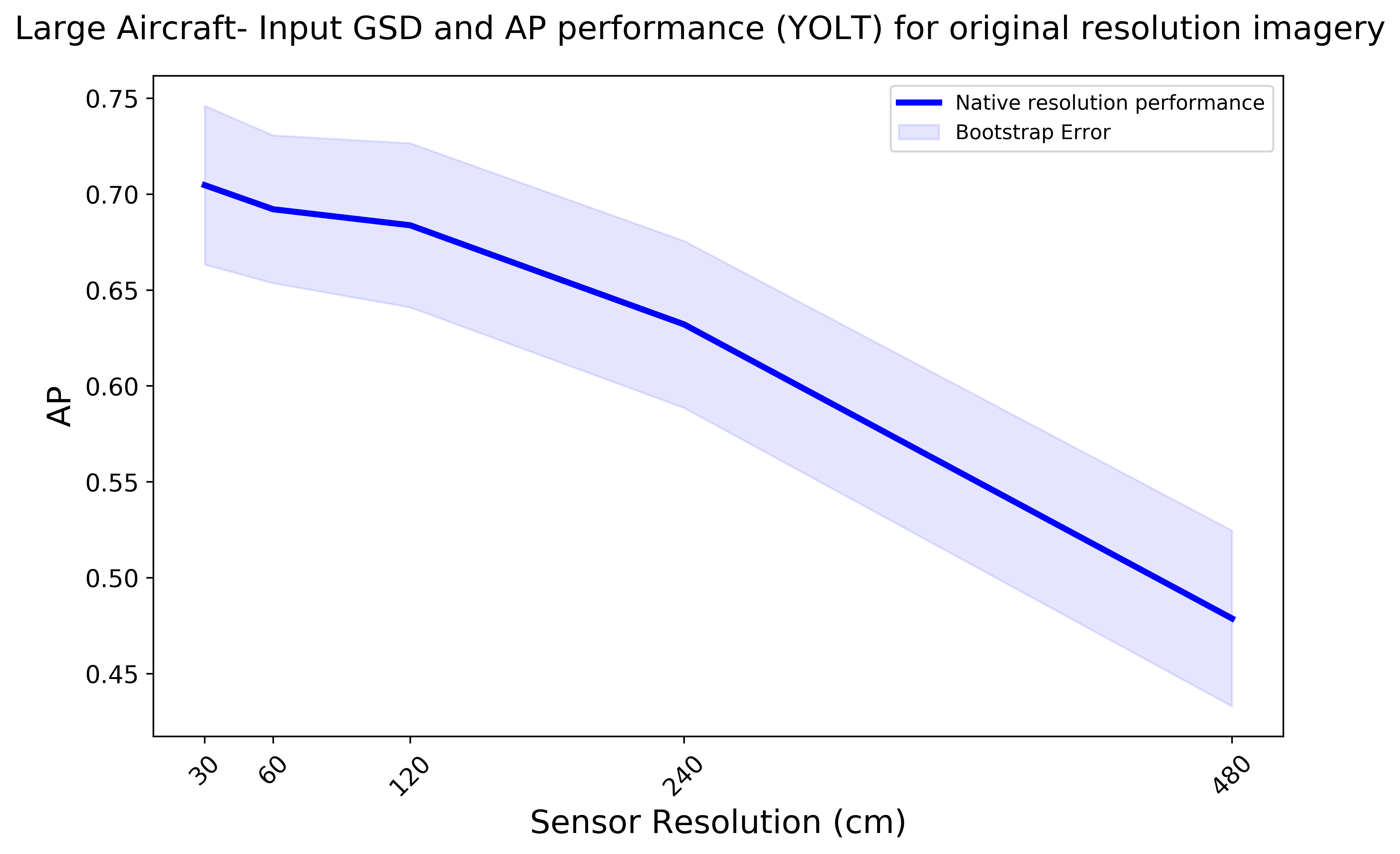}
 \caption{Large Aircraft- Native Resolution Performance }
    \label{fig:1}
  \end{subfigure}
	\hfill %
  \begin{subfigure}[b]{0.48\columnwidth}
    \includegraphics[width=\linewidth]{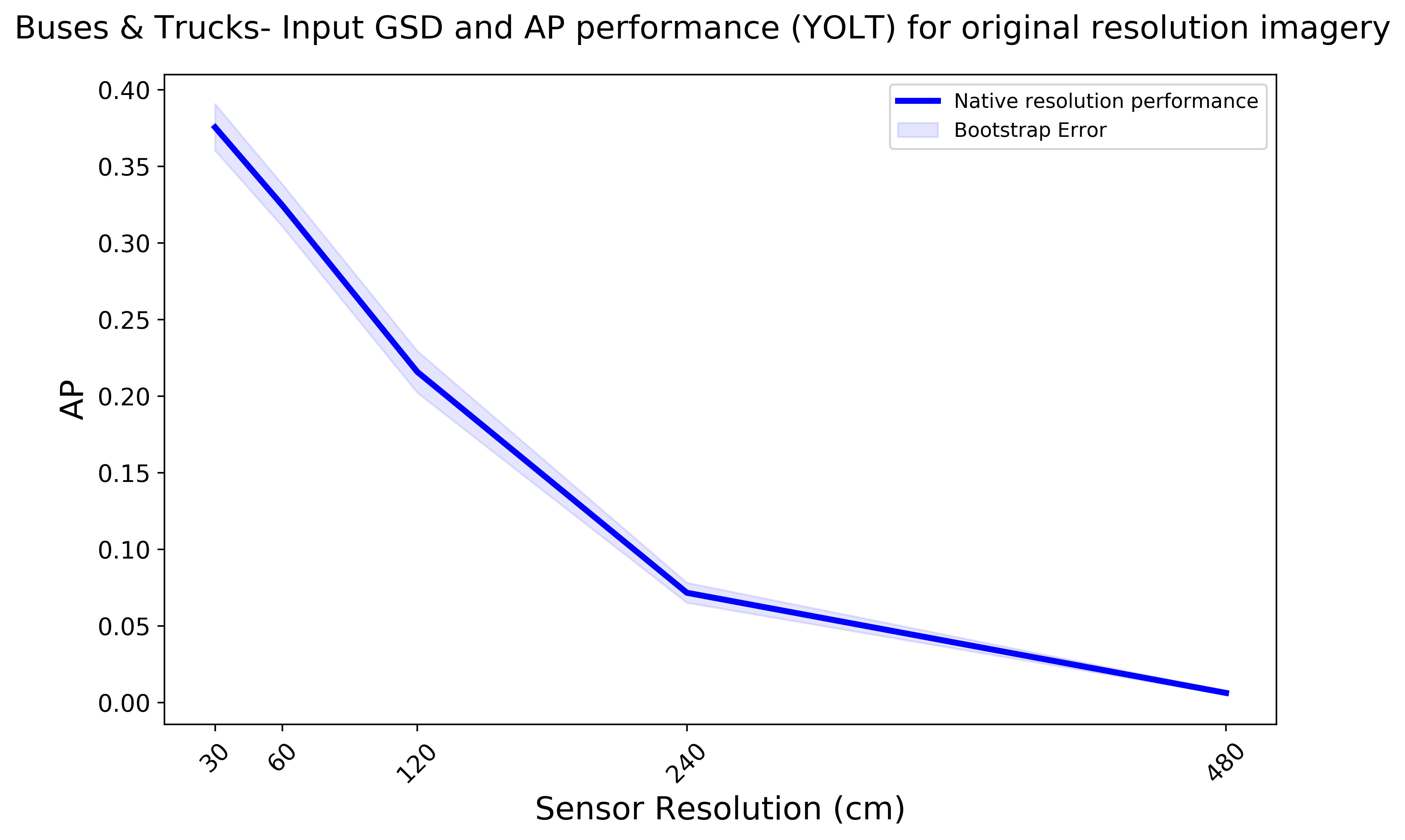}
  \caption{Buses/Trucks- Native Resolution Performance }
    \label{fig:2}
  \end{subfigure}
 	 \hfill %
\vspace{3pt}
 \begin{subfigure}[b]{0.48\columnwidth}
    \includegraphics[width=\linewidth]{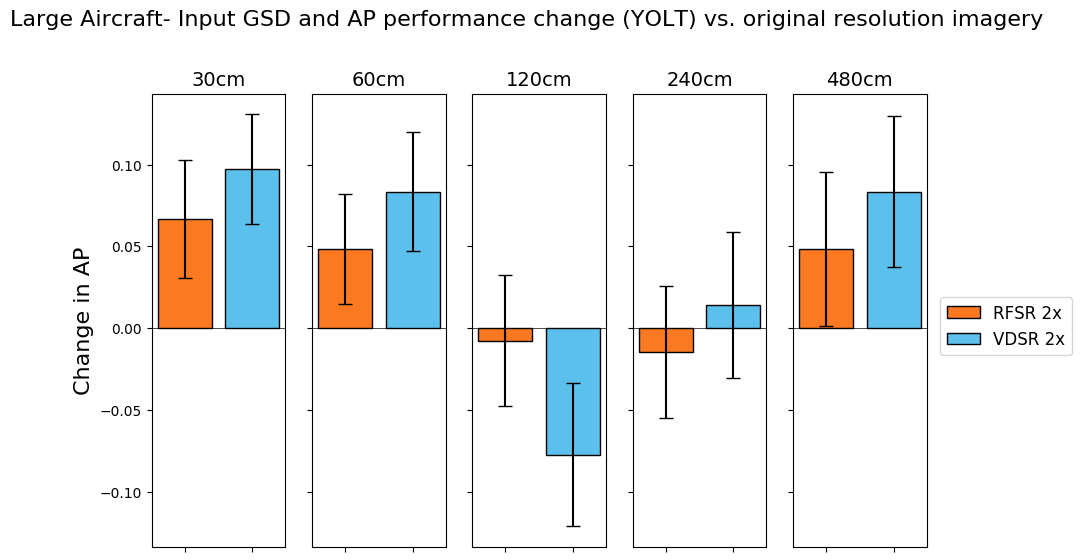}
 \caption{Large Aircraft - AP Change}
    \label{fig:1}
  \end{subfigure}
	\hfill %
  \begin{subfigure}[b]{0.48\columnwidth}
    \includegraphics[width=\linewidth]{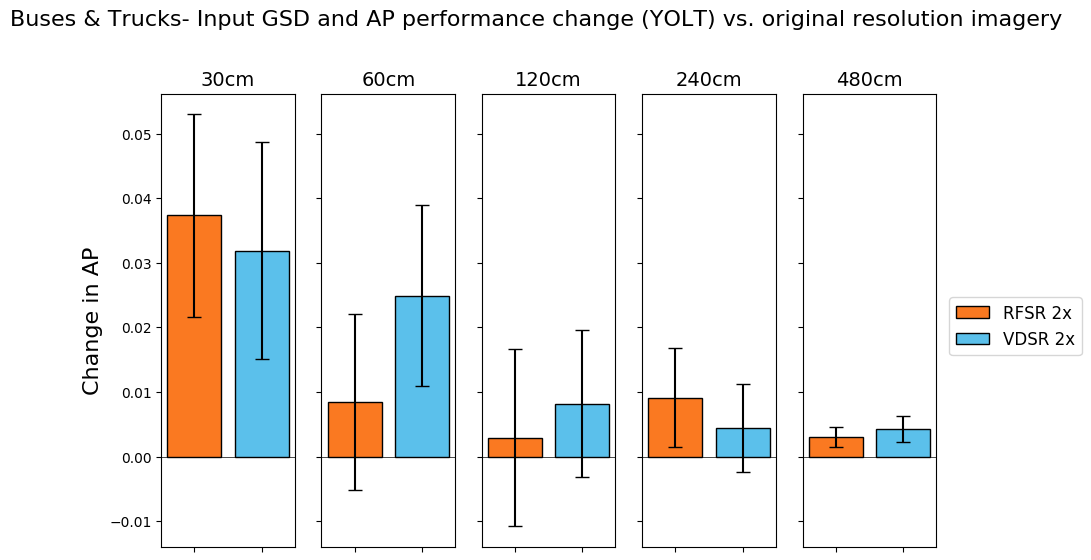}
  \caption{Buses/Trucks - AP Change}
    \label{fig:2}
  \end{subfigure}
 	 \hfill %
 \vspace{-2mm}%Put here to reduce too much white space after your table 
 \caption{YOLT performance curves for Large Aircraft (a) and Buses/Trucks (b) at native resolution. Performance change versus these curves when super-enhancing imagery $2\times$ (c and d).}
 \label{fig:sl1}
 \vspace{-4mm}%Put here to reduce too much white space after your table 
\end{figure}

Furthermore, although performance is much worse with SSD, super-resolution techniques are much more effective. With SSD, for both RFSR and VDSR performance boosts are evident for all resolutions, except for 60 cm to 30 cm.  VDSR is generally shown to be slightly superior to RFSR when detecting objects with SSD.   For SSD the improvement at 480 cm is statistically quite significant, though this is primarily due to the mAP of 0.0 for native imagery.   Performance increases significantly once objects are greater than $\approx20$ pixels in extent. This trend extends across object classes, as shown in the performance curves for individual object classes (See Supplemental Material). %Also apparent in Figure \ref{fig:sl1}b is that a smaller field of view is preferred for detection of densely packed object such as trucks.

\section{Conclusions}
%What do we actually learn as a result of this study?  What is novel?
%Discuss statistical significance of increase (best at 30 cm)
%Discuss performance by object class, and as a function of  number of pixels.
%Include curves for each category in supplemental, and more object detection outputs at various resolutions (also SSD).
%Compare to SIMRDWN results:
%	SSD: Airplane=0.56, Boat=0.44, Car=0.57
%	YOLT: Airplane=0.75, Boat=0.68, Car=0.71

In this paper we undertook a rigorous study of the utility provided by super-resolution techniques towards the detection of objects in satellite imagery.  We paired two super-resolution techniques (VDSR and RFSR) with advanced object detection methods and searched for objects in a satellite imagery dataset with over 250,000 labeled objects in a diverse set of environments.  In order to establish super-resolution effects at multiple sensor resolutions, we degrade this imagery from 30 cm to 60, 120, 240, and 480 cm resolutions.  Our baseline tests with both the YOLT and SSD models of the SIMRDWN object detection framework indicate that object detection performance decreases by $22 - 27\%$ when resolution degrades from 30 cm to 120 cm.  

The application of SR techniques as a pre-processing step provides an improvement in object detection performance at most resolutions (Table \ref{tab:det_perf}). For both object detection frameworks, the greatest benefit is achieved at the highest resolutions, as super-resolving native 30 cm imagery to 15 cm yields a $13 - 36\%$ improvement in mAP. Furthermore, when using YOLT, we find that enhancing imagery from 60 cm to 15 cm provides a significant boost in performance over both the native 30 cm imagery ($+13\%$) and native 60 cm imagery ($+20\%$). The performance boost applies to all classes, but is most significant for boats, large aircraft, and buses/trucks. Again with YOLT, in coarser resolutions (120 cm to 480 cm) SR provides little to no boost in performance (-0.02 to +0.04 change in mAP). When using SSD, super-resolving imagery from 30 to 15 cm provides a substantial boost for the identification of small vehicles ($+56\%$), but provides mixed results for other classes (See supplemental material).  In coarser resolutions, with SSD, SR techniques provide a greater boost in performance however the performance for most classes is still worse compared to YOLT with native imagery.  Overall, given the relative ease of applying SR techniques, the general improvement observed in this study is noteworthy and suggests SR could be a valuable pre-processing step for future object detection applications with satellite imagery.

{\small
\bibliographystyle{ieee}
\bibliography{main.bib}
}

%%%%%%%%%%%%%%%% SUPPLEMENTARY

\pagebreak
\onecolumn
\begin{center}
\Large{
  \textbf{The Effects of Super-Resolution on Object Detection Performance in Satellite Imagery (Supplemental Material)}}\\[0.5cm]
\end{center}
\setcounter{equation}{0}
\setcounter{figure}{0}
\setcounter{table}{0}
\setcounter{section}{0}
\setcounter{page}{1}
\makeatletter
\renewcommand{\theequation}{S\arabic{equation}}
\renewcommand{\thefigure}{S\arabic{figure}}

\section{Super-Resolution Outputs}

\begin{figure}[h]
\begin{center}
\includegraphics[width=0.9\linewidth]{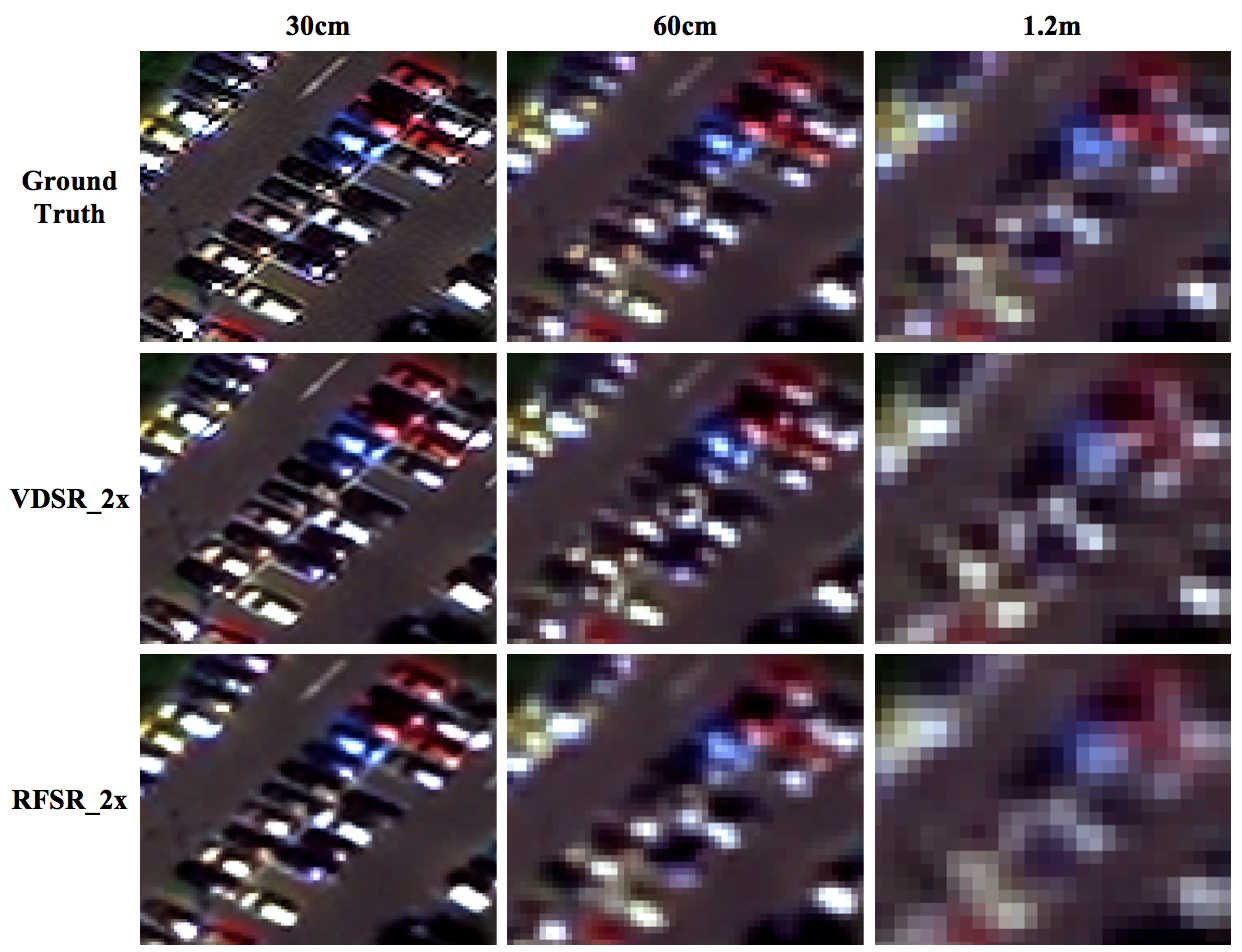}
\end{center}
\vspace{-4mm}%Put here to reduce too much white space after your table 
\caption{Super-resolution example as applied to images containing cars. Smaller objects like cars rapidly degrade and become amorphous pixelated blobs as resolution degrades. For the 30 cm data and super-resolved outputs, VDSR and RFSR perform favorably using a $2\times$ enhancement (60 cm input $\rightarrow$ 30 cm SR output). VDSR can generally recover the car shape of the 30 cm data with a $4\times$ enhancement (120 cm input $\rightarrow$ 30 cm SR output), whereas RFSR struggles to do so. 
}
\label{fig:sn_fig1}
%\vspace{-4mm}%Put here to reduce too much white space after your table 
\end{figure}

\begin{figure}[h]
\begin{center}
\includegraphics[width=0.9\linewidth]{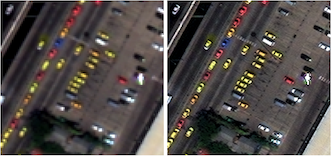}
\end{center}
\vspace{-4mm}%Put here to reduce too much white space after your table 
\caption{Ground truth 30 cm imagery (right), and simulated 60 cm input imagery (left).  
}
\label{fig:sn_fig1}
%\vspace{-4mm}%Put here to reduce too much white space after your table 
\end{figure}

\begin{figure}[h]
\begin{center}
\includegraphics[width=0.9\linewidth]{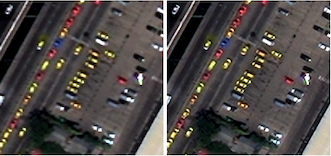}
\end{center}
\vspace{-4mm}%Put here to reduce too much white space after your table 
\caption{Super-resolution with the $60\rightarrow 30$ models (left = $2\times$RFSR, right = 2$\times$VDSR).
%with the RFSR $2\times$ model  30 cm. 
}
\label{fig:sn_fig1}
%\vspace{-4mm}%Put here to reduce too much white space after your table 
\end{figure}

\newpage
\clearpage

\section{Super-Resolution Scores- Bicubic Decimation with No Blurring}

\begin{table}[h]
%\resizebox{\textwidth}{!}{%
\footnotesize%\small%\tiny
\begin{center}
\begin{tabular}{lllll}
\hline
GSD$_{\rm out}$ & Scale &     Bicubic &    VDSR &    RFSR \\
\hline\hline
30cm       & ×2    & 40.30 / 0.8734 & 42.95 / 0.9104 & 40.90 / 0.8885 \\
30cm       & ×4    & 36.83 / 0.7265 & 39.07 / 0.7939 & 36.81 / 0.7419 \\
30cm       & ×8    & 34.69 / 0.5930 & 36.86 / 0.6605 & 35.06 / 0.6092 \\
\hline
60cm       & ×2    & 43.69 / 0.9594 & 45.66 / 0.9717 & 44.25 / 0.9496 \\
60cm       & ×4    & 38.61 / 0.8656 & 41.07 / 0.9055 & 38.89 / 0.8729 \\
60cm       & ×8    & 35.20 / 0.7324 & 37.61 / 0.7861 & 35.18 / 0.7439 \\
\hline
1.2m       & ×2    & 40.73 / 0.9640 & 43.14 / 0.9777 & 41.70 / 0.9716 \\
1.2m       & ×4    & 34.73 / 0.8562 & 37.39 / 0.9006 & 35.23 / 0.8773 \\
1.2m       & ×8    & 30.92 / 0.6862 & 33.16 / 0.7500 & 31.19 / 0.7197 \\
\hline
2.4m       & ×2    & 39.33 / 0.9529 & 42.15 / 0.9720 & 40.36 / 0.9640 \\
2.4m       & ×4    & 33.24 / 0.7998 & 36.00 / 0.8638 & 33.74 / 0.8326 \\
2.4m       & ×8    & 29.68 / 0.5802 & 31.90 / 0.6566 & 29.92 / 0.6177 \\
\hline
4.8m       & ×2    & 37.99 / 0.9320 & 40.94 / 0.9592 & 38.98 / 0.9491 \\
4.8m       & ×4    & 32.23 / 0.7294 & 34.87 / 0.8070 & 32.66 / 0.7703 \\
4.8m       & ×8    & 29.01 / 0.4786 & 31.15 / 0.5552 & 29.19 / 0.5131 \\
\hline
\end{tabular}%}
\end{center}
\vspace{-4mm}%Put here to reduce too much white space after your table 
\caption{Average PSNR / SSIM scores for scale  $2\times$,  $4\times$,  and  $8\times$ across five super-resolution output GSDs. All test imagery is not blurred, decimated bicubically, and then bicubically upsampled.  (Ideal or Traditional super-resolution settings).  As in the main text the xView validation dataset is used (281 images).  Bicubic indicates the scores if LR images are just upscaled using bicubic interpolation to match the HR image size.}
\label{tab:sr_perf}
\vspace{-4mm}%Put here to reduce too much white space after your table 
\end{table}

\newpage
\clearpage

\section{Object Detection Performance}

The figures below illustrate bounding boxes output by YOLT models at various resolutions.  Cars are green, buses/trucks are blue, small aircraft are red, and large aircraft are yellow.

\begin{figure}[h!]
\begin{center}
\includegraphics[width=0.65\linewidth]{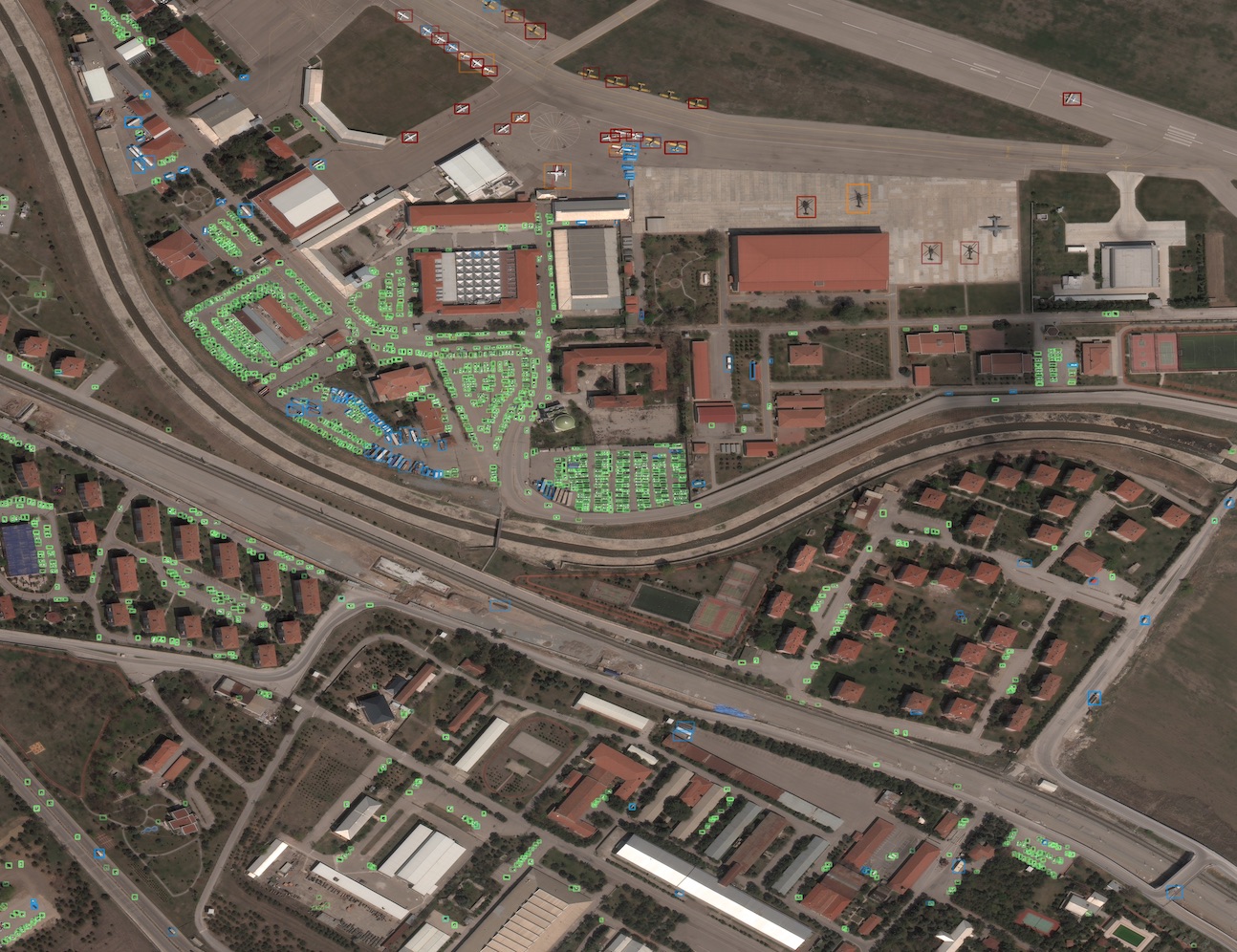}
\end{center}
\vspace{-4mm}%Put here to reduce too much white space after your table 
\caption{Performance of YOLT model trained and tested on native 30 cm imagery for image\_id=1114 and a low detection threshold of 0.1 (this detection threshold yields fewer false negatives but more false positives). }
\label{fig:z0}
\end{figure}

\begin{figure}[b]
\begin{center}
\includegraphics[width=0.65\linewidth]{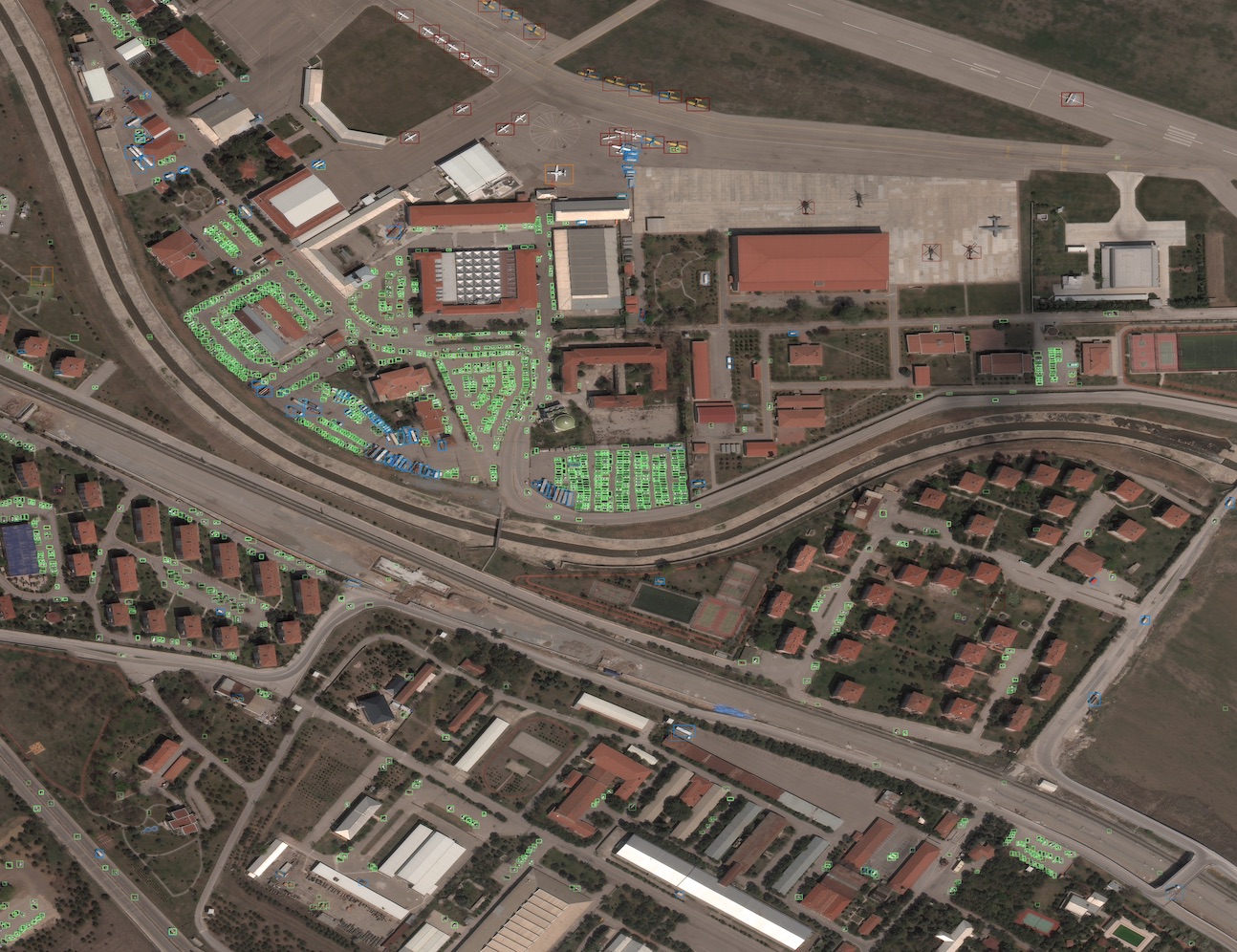}
\end{center}
\vspace{-4mm}%Put here to reduce too much white space after your table 
\caption{Performance of YOLT model trained and tested on super-resolved 15 cm imagery from the VDSR $2\times$ model. We use a low detection threshold of 0.1 (this detection threshold yields fewer false negatives but more false positives).}
\label{fig:z0}
\end{figure}

\begin{figure}[]
\begin{center}
\includegraphics[width=0.65\linewidth]{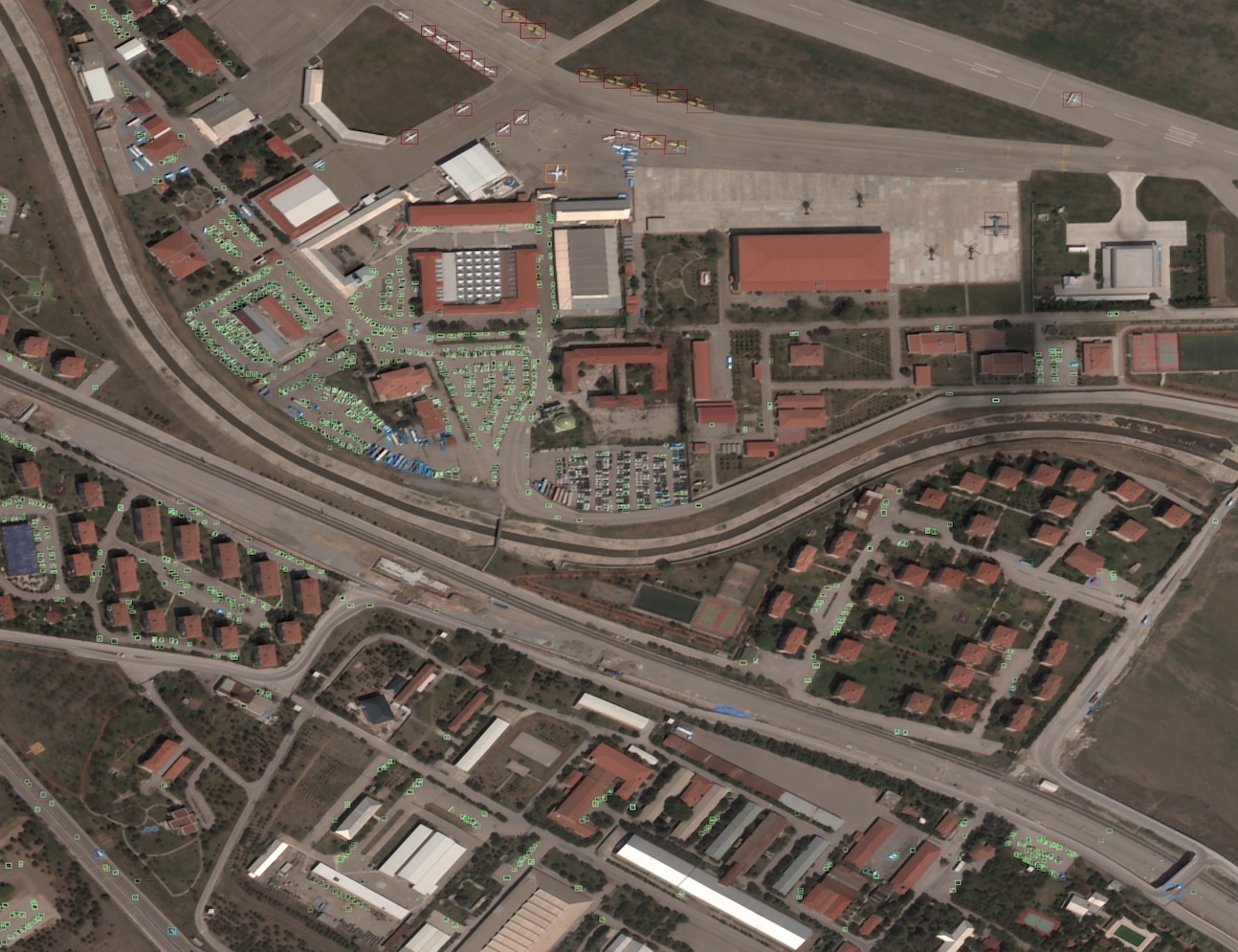}
\end{center}
\vspace{-4mm}%Put here to reduce too much white space after your table 
\caption{Performance of YOLT model trained and tested on super-resolved 30 cm imagery from the RFSR $4\times$ model. We use a low detection threshold of 0.1 (this detection threshold yields fewer false negatives but more false positives).}
\label{fig:z0}
\end{figure}

\begin{figure}[]
\begin{center}
\includegraphics[width=0.65\linewidth]{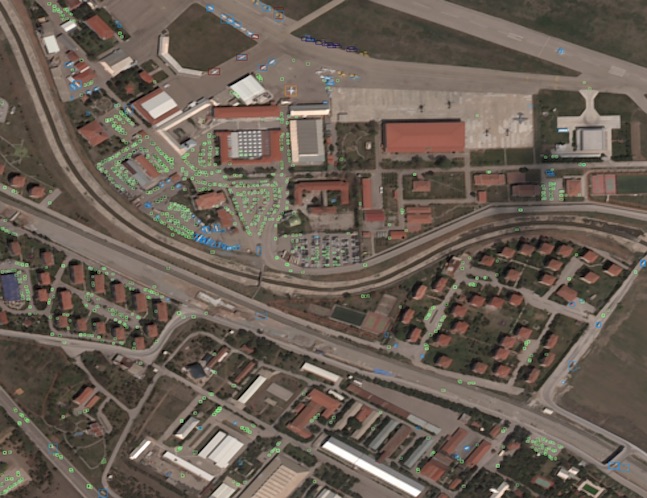}
\end{center}
\vspace{-4mm}%Put here to reduce too much white space after your table 
\caption{Performance of YOLT model trained and tested on super-resolved 60 cm imagery from the RFSR $4\times$ model. We use a low detection threshold of 0.1 (this detection threshold yields fewer false negatives but more false positives).}
\label{fig:z0}
\end{figure}

\begin{figure}[t!]
\begin{center}
\includegraphics[width=0.65\linewidth]{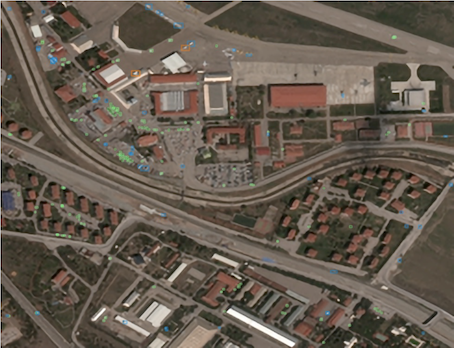}
\end{center}
\vspace{-4mm}%Put here to reduce too much white space after your table 
\caption{Performance of YOLT model trained and tested on super-resolved 120 cm imagery from the VDSR $4\times$ model. We use a low detection threshold of 0.1 (this detection threshold yields fewer false negatives but more false positives).}
\label{fig:z0}
\end{figure}

%%%%%%%%%%%%%%%%%%%
\clearpage
\newpage
\section{Object Detection Performance Curves and Tables}

To compute the statistical difference ($\sigma_{diff}$) between the super-resolved and baseline models, we follow the procedure used in Table 4 of the main text. The joint error estimate between the baseline ($Y_b$) and super-resolved ($Y_{sr}$) data point can be estimated as:

\begin{equation}
\sigma_{tot}^2 = \sigma_b^2 + \sigma_{sr}^2
\label{eqn:sigma_tot}
\end{equation}

The statistical difference between two models is then simply:

\begin{equation}
\sigma_{diff} = \frac{Y_{sr} - Y_b}{\sigma_{tot}}
\label{eqn:sigma_diff}
\end{equation}

The tables and plots below show the performance of various models for each object class.
\clearpage
\newpage

\begin{figure}[t]
\vspace{-0mm}%Put here to reduce too much white space after your table 
\begin{center}
\includegraphics[width=0.5\linewidth]{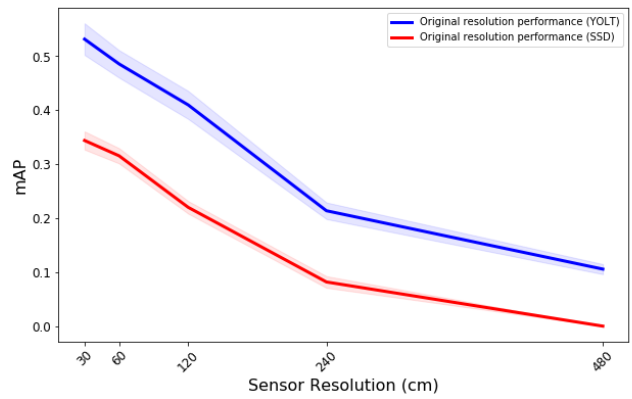}
\end{center}
\vspace{-6mm}%Put here to reduce too much white space after your table 
\caption{Performance of YOLT and SSD at the native sensor resolution for all object classes.}
\label{fig:S_yolt_ssd_curve}
%\vspace{-4mm}%Put here to reduce too much white space after your table 
\end{figure}

\begin{figure}[t]
\vspace{-0mm}%Put here to reduce too much white space after your table 
\begin{center}
\includegraphics[width=0.666\linewidth]{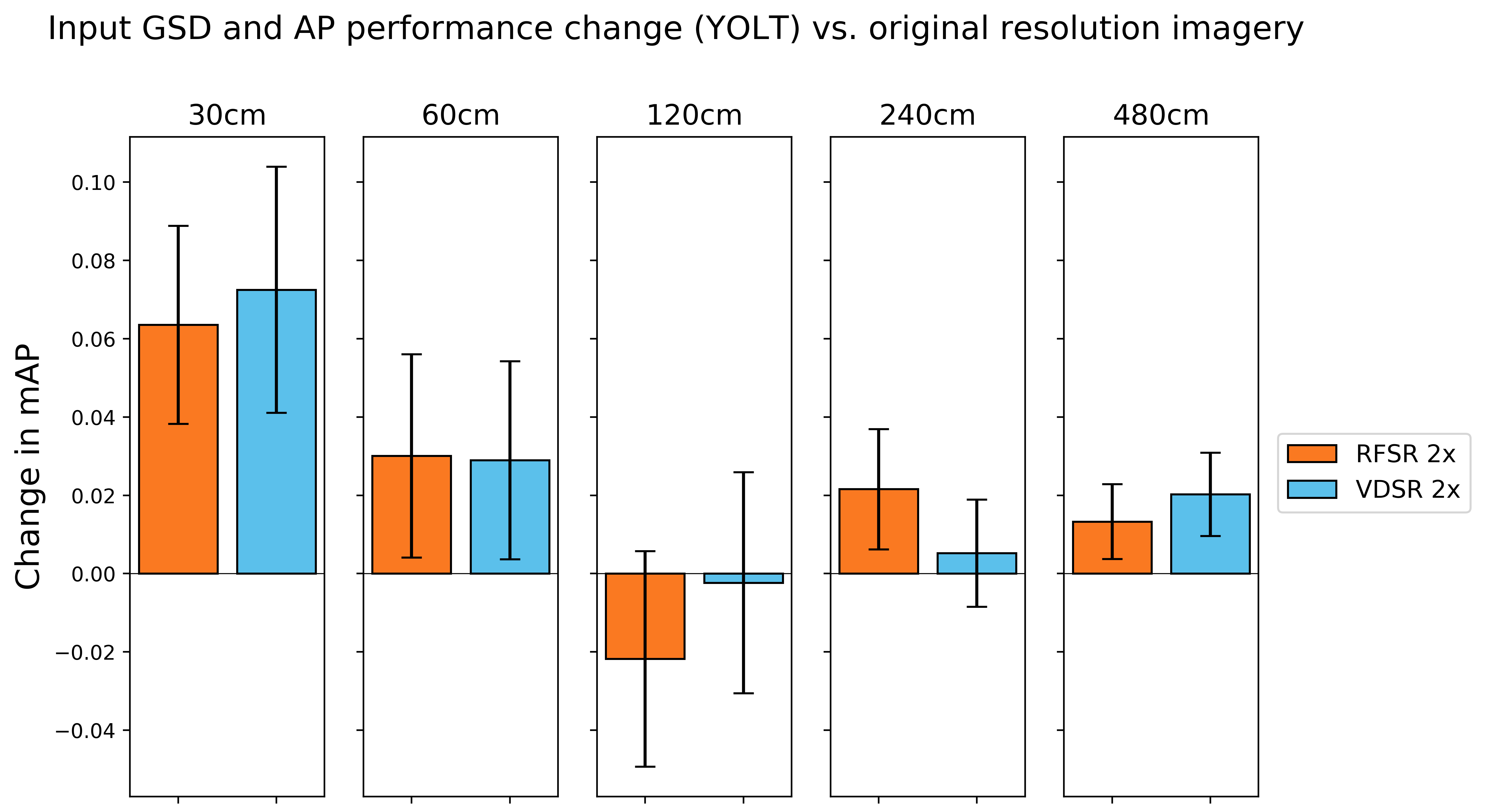}
\end{center}
\vspace{-6mm}%Put here to reduce too much white space after your table 
\caption{Performance change over original resolution (Figure \ref{fig:S_yolt_ssd_curve}-Blue Line) using YOLT and 2x super-resolved data.}
\label{fig:yolt_bars}
\vspace{-4mm}%Put here to reduce too much white space after your table 
\end{figure}

\begin{figure}[t]
\vspace{-0mm}%Put here to reduce too much white space after your table 
\begin{center}
\includegraphics[width=0.666\linewidth]{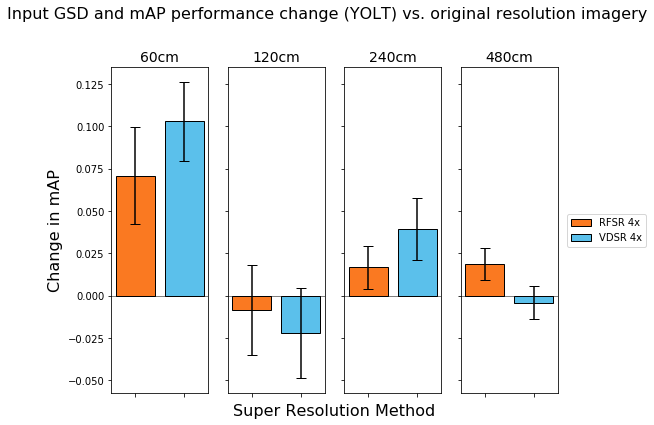}
\end{center}
\vspace{-6mm}%Put here to reduce too much white space after your table 
\caption{Performance change over original resolution (Figure \ref{fig:S_yolt_ssd_curve}-Blue Line) using YOLT and 4x super-resolved data.}
\label{fig:yolt_bars}
\vspace{-4mm}%Put here to reduce too much white space after your table 
\end{figure}

\begin{table*}[t]
%\resizebox{\textwidth}{!}{%
\scriptsize%\footnotesize%\small%\tiny
\begin{center}
\begin{tabular}{lllllllll}
\hline
Model & Data           & 30 cm                          & 60 cm                          & 120 cm                         & 240 cm                         & 480 cm                         \\ \hline\hline
YOLT  & Native         & $0.53 \pm 0.03$                & $0.49 \pm 0.03$                & $0.41 \pm 0.03$                & $0.21 \pm 0.02$                & $0.11 \pm 0.01$                \\
YOLT  & RFSR $2\times$ & $0.60 \pm 0.03$ (+1.6$\sigma$) & $0.52 \pm 0.03$ (+0.8$\sigma$) & $0.39 \pm 0.03$ (-0.6$\sigma$) & $0.24 \pm 0.02$ (+1.0$\sigma$) & $0.12 \pm 0.01$ (+1.0$\sigma$) \\
YOLT  & VDSR $2\times$ & $0.60 \pm 0.03$ (+1.7$\sigma$) & $0.52 \pm 0.03$ (+0.8$\sigma$) & $0.41 \pm 0.03$ (-0.1$\sigma$) & $0.22 \pm 0.01$ (+0.3$\sigma$) & $0.13 \pm 0.01$ (+1.4$\sigma$) \\
YOLT  & RFSR $4\times$ &                                & $0.56 \pm 0.03$ (+1.9$\sigma$) & $0.40 \pm 0.03$ (-0.2$\sigma$) & $0.23\pm0.01$ (+0.8$\sigma$)   & $0.12\pm0.01$ (+1.4$\sigma$)   \\
YOLT  & VDSR $4\times$ &                                & $0.59 \pm 0.02$ (+3.0$\sigma$) & $0.39 \pm 0.03$ (-0.6$\sigma$) & $0.25\pm0.02$ (+1.7$\sigma$)   & $0.10\pm0.01$ (-0.3$\sigma$)   \\ \hline
SSD   & Native         & $0.30\pm0.01$                  & $0.32\pm0.01$                  & $0.22\pm0.01$                  & $0.08\pm0.01$                  & $0.00 \pm 0.00$                \\
SSD   & RFSR $2\times$ & $0.36\pm0.01$ (+2.6$\sigma$)   & $0.33\pm0.02$ (+0.7$\sigma$)   & $0.24\pm0.01$ (+1.1$\sigma$)   & $0.13\pm0.01$ (+3.2$\sigma$)   & $0.07\pm0.01$ (+7.0$\sigma$)   \\
SSD   & VDSR $2\times$ & $0.41\pm0.03$ (+3.5$\sigma$)   & $0.32\pm0.02$ (-0.0$\sigma$)   & $0.26\pm0.01$ (+2.3$\sigma$)   & $0.14\pm0.01$ (+3.9$\sigma$)   & $0.08\pm0.01$ (+8.7$\sigma$)   \\ \hline
\end{tabular}%}
\end{center}
\vspace{-4mm}%Put here to reduce too much white space after your table 
\caption{Performance for each data type in mAP.  For RFSR and VDSR at each resolution we note the error and statistical difference from the baseline model (e.g. +0.5$\sigma$). The native sensor resolution of our original imagery and the input into the super-resolution models  is shown on the X-axis.  We then compare the super-resolved outputs vs. the original native imagery to test the change in object detection performance. }
\label{tab:det_perf}
\vspace{-4mm}%Put here to reduce too much white space after your table 
\end{table*}

\clearpage
\newpage

\begin{figure}[h!]
\begin{center}
\includegraphics[width=0.7\linewidth]{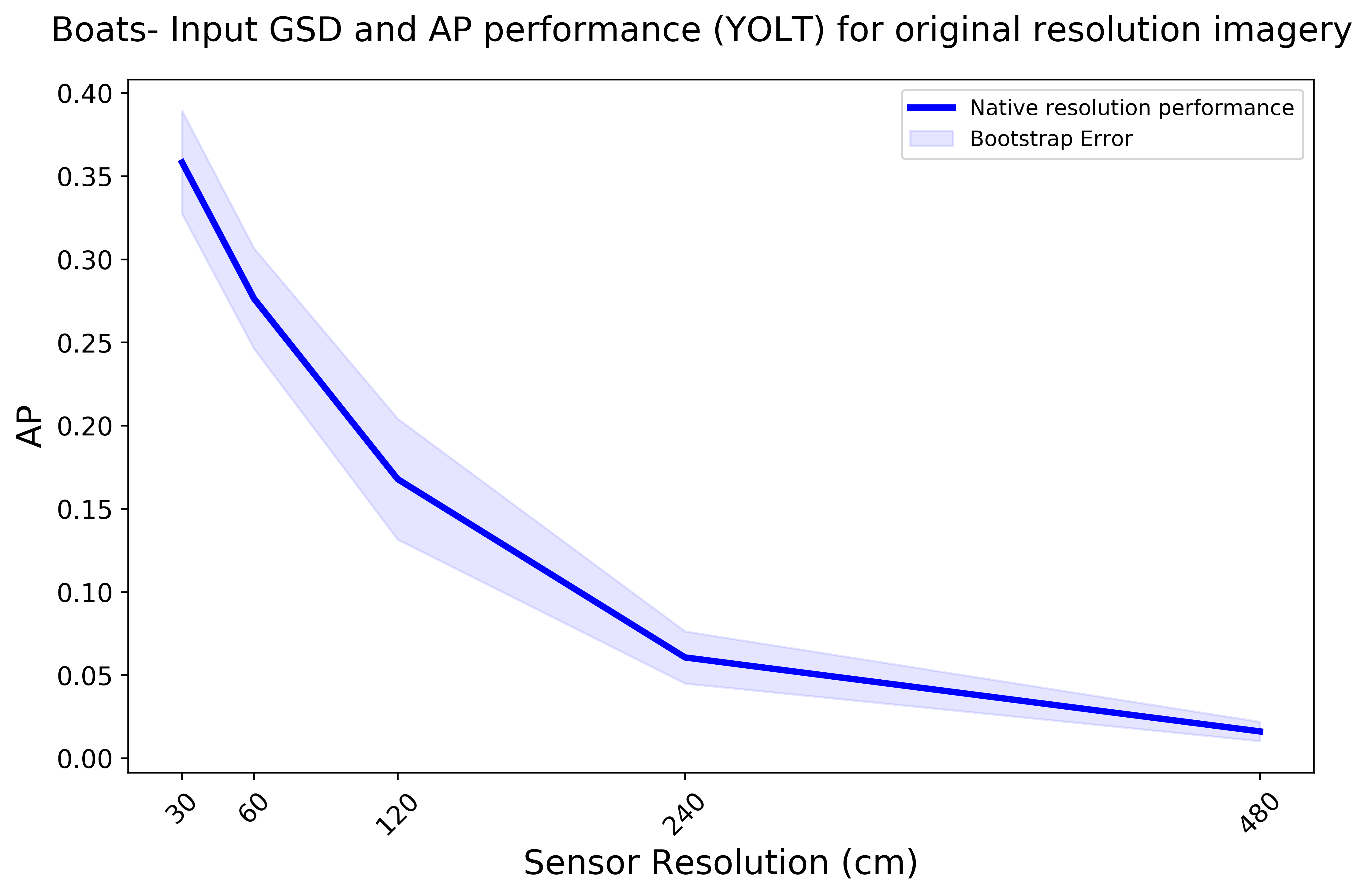}
\end{center}
\vspace{-4mm}%Put here to reduce too much white space after your table 
\caption{Performance of YOLT models on boats as a function of sensor resolution.  The lower axis indicates the sensor resolution, with average precision plotted on the y-axis.}
\label{fig:boats}
\end{figure}

\begin{figure}[h!]
\begin{center}
\includegraphics[width=0.7\linewidth]{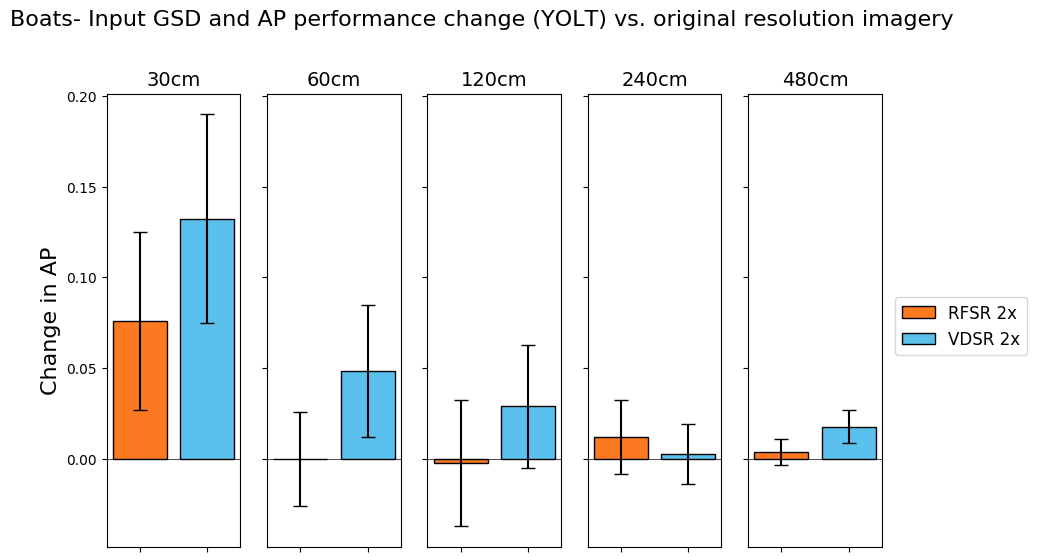}
\end{center}
\vspace{-4mm}%Put here to reduce too much white space after your table 
\caption{Performance change over original resolution (Figure \ref{fig:boats} using YOLT and 2x super-resolved data (Boats)).}
\end{figure}

\begin{figure}[h!]
\begin{center}
\includegraphics[width=0.7\linewidth]{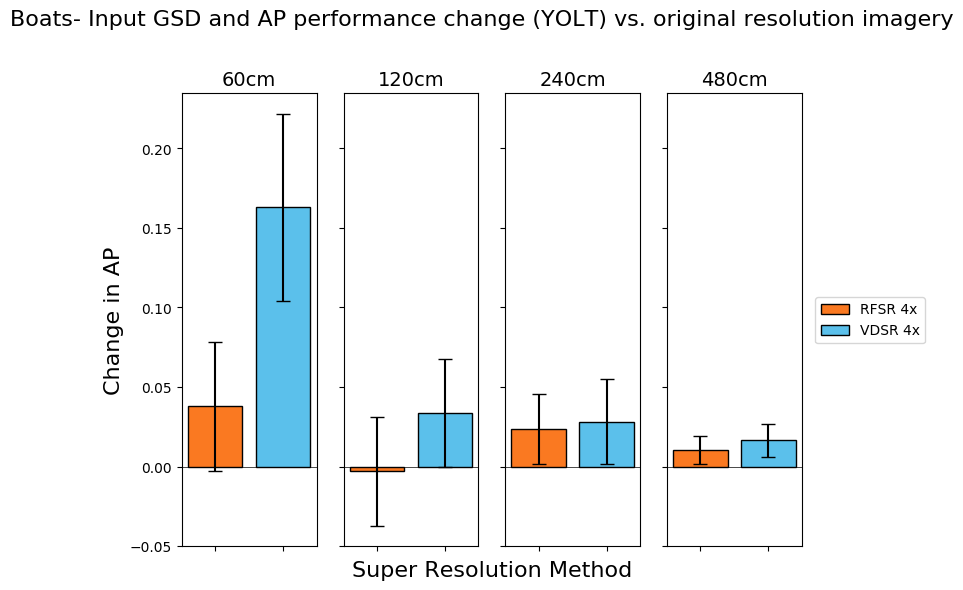}
\end{center}
\vspace{-4mm}%Put here to reduce too much white space after your table 
\caption{Performance change over original resolution (Figure \ref{fig:boats} using YOLT and 4x super-resolved data (Boats)).}
\end{figure}

\begin{table*}[h!]
%\resizebox{\textwidth}{!}{%
\scriptsize%\footnotesize%\small%\tiny
\begin{center}
\begin{tabular}{lllllllll}
\hline
Model & Data           & 30 cm                          & 60 cm                          & 120 cm                         & 240 cm                         & 480 cm                         \\ \hline\hline
YOLT  & Native         & $0.36 \pm 0.03$                & $0.28 \pm 0.03$                & $0.17 \pm 0.04$                & $0.06 \pm 0.02$                & $0.02 \pm 0.01$                \\
YOLT  & RFSR $2\times$ & $0.43 \pm 0.05$ (+1.3$\sigma$) & $0.28 \pm 0.03$ (+0.0$\sigma$) & $0.17 \pm 0.03$ (+0.0$\sigma$) & $0.07 \pm 0.02$ (+0.5$\sigma$) & $0.02 \pm 0.01$ (+0.4$\sigma$) \\
YOLT  & VDSR $2\times$ & $0.49 \pm 0.03$ (+2.0$\sigma$) & $0.32 \pm 0.03$ (+1.0$\sigma$) & $0.20 \pm 0.03$ (+0.6$\sigma$) & $0.06 \pm 0.01$ (+0.1$\sigma$) & $0.03 \pm 0.01$ (+1.6$\sigma$) \\
YOLT  & RFSR $4\times$ &                                & $0.31 \pm 0.04$ (+0.7$\sigma$) & $0.16 \pm 0.03$ (-0.1$\sigma$) & $0.08\pm0.02$ (+0.9$\sigma$)   & $0.03\pm0.01$ (+1.0$\sigma$)   \\
YOLT  & VDSR $4\times$ &                                & $0.44 \pm 0.06$ (+2.5$\sigma$) & $0.20 \pm 0.03$ (+0.7$\sigma$) & $0.09\pm0.03$ (+0.9$\sigma$)   & $0.03\pm0.01$ (+1.4$\sigma$)   \\ 
\hline
SSD   & Native         & 0.12 $\pm$ 0.02                & 0.14 $\pm$ 0.03                & 0.03 $\pm$ 0.01                & 0.01 $\pm$ 0                   & 0 $\pm$ 0                      \\
SSD   & RFSR $2\times$ & 0.14 $\pm$ 0.03 (+0.6$\sigma$) & 0.09 $\pm$ 0.02 (-1.3$\sigma$) & 0.05 $\pm$ 0.01 (+0.9$\sigma$) & 0.02 $\pm$ 0.01 (+1.1$\sigma$) & 0.01 $\pm$ 0 (+1.8$\sigma$)    \\
SSD   & VDSR $2\times$ & 0.11 $\pm$ 0.02 (-0.2$\sigma$) & 0.1 $\pm$ 0.02 (-0.9$\sigma$)  & 0.04 $\pm$ 0.01 (-2.9$\sigma$) & 0.03 $\pm$ 0.01 (+2.8$\sigma$) & 0.01 $\pm$ 0 (+2.1$\sigma$)   \\
\hline
\end{tabular}%}
\end{center}
\vspace{-4mm}%Put here to reduce too much white space after your table 
\caption{Performance for the boat class.  For RFSR and VDSR at each resolution we note the error and statistical difference from the baseline model (e.g. +0.5$\sigma$). }
\label{tab:det_perf}
\end{table*}

\newpage
\clearpage

\begin{figure}[h!]
\begin{center}
\includegraphics[width=0.7\linewidth]{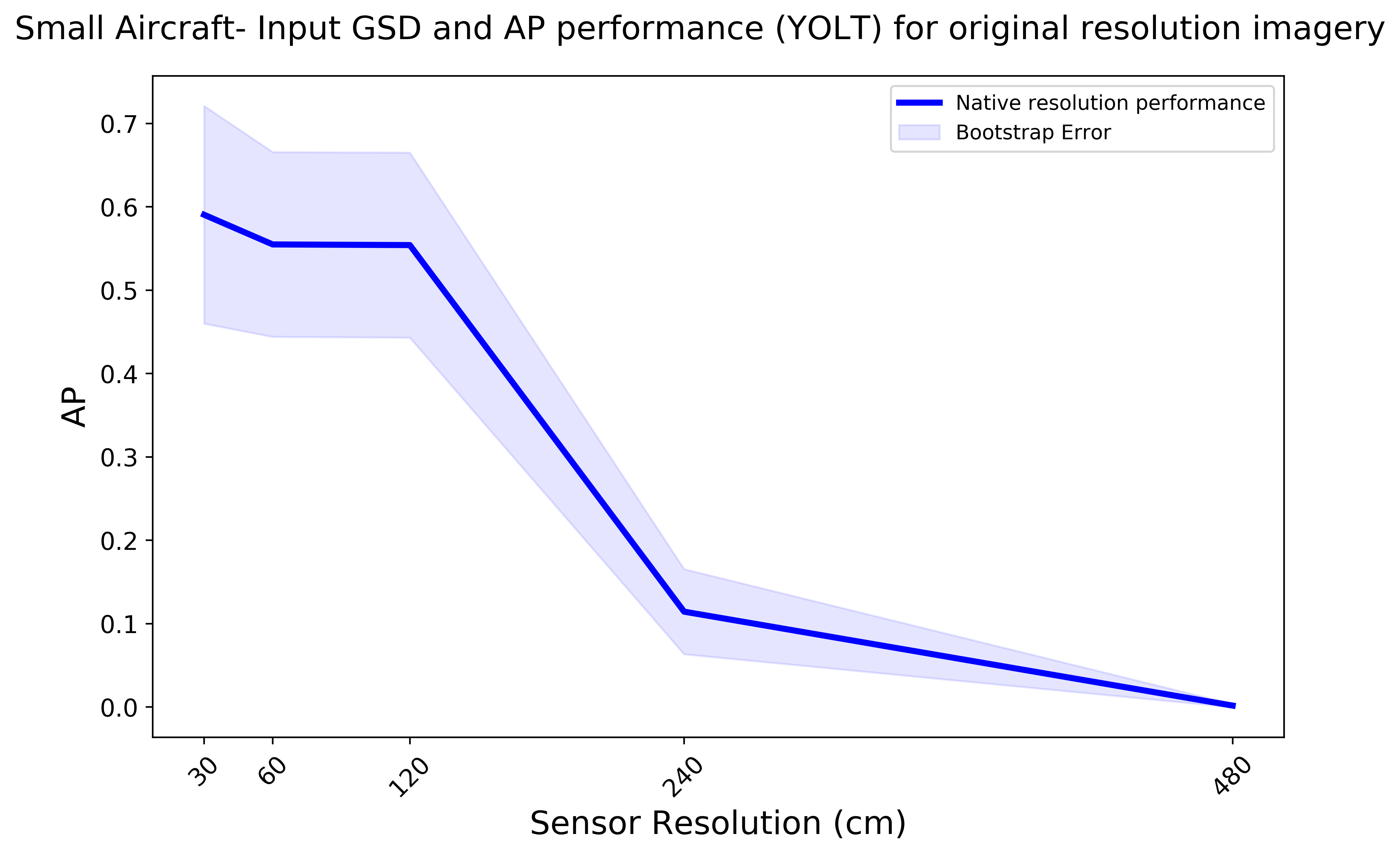}
\end{center}
\vspace{-4mm}%Put here to reduce too much white space after your table 
\caption{Performance of YOLT models on small aircraft as a function of sensor resolution.  The lower axis indicates the sensor resolution, with average precision plotted on the y-axis.}
\label{fig:sair}
\end{figure}

\begin{figure}[h!]
\begin{center}
\includegraphics[width=0.7\linewidth]{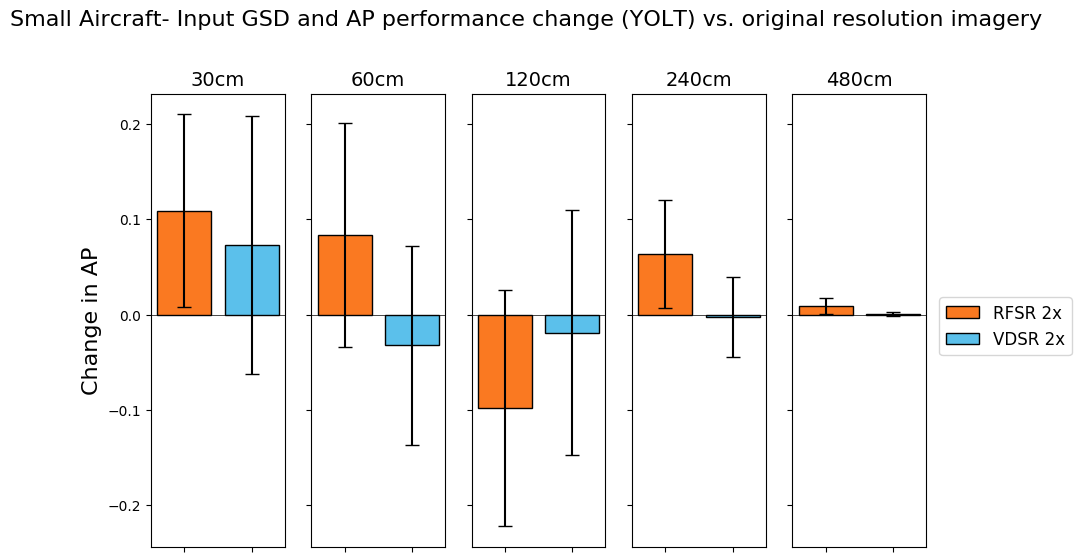}
\end{center}
\vspace{-4mm}%Put here to reduce too much white space after your table 
\caption{Performance change over original resolution (Figure \ref{fig:sair} using YOLT and 2x super-resolved data (Small Aircraft)).}
\end{figure}

\begin{figure}[h!]
\begin{center}
\includegraphics[width=0.7\linewidth]{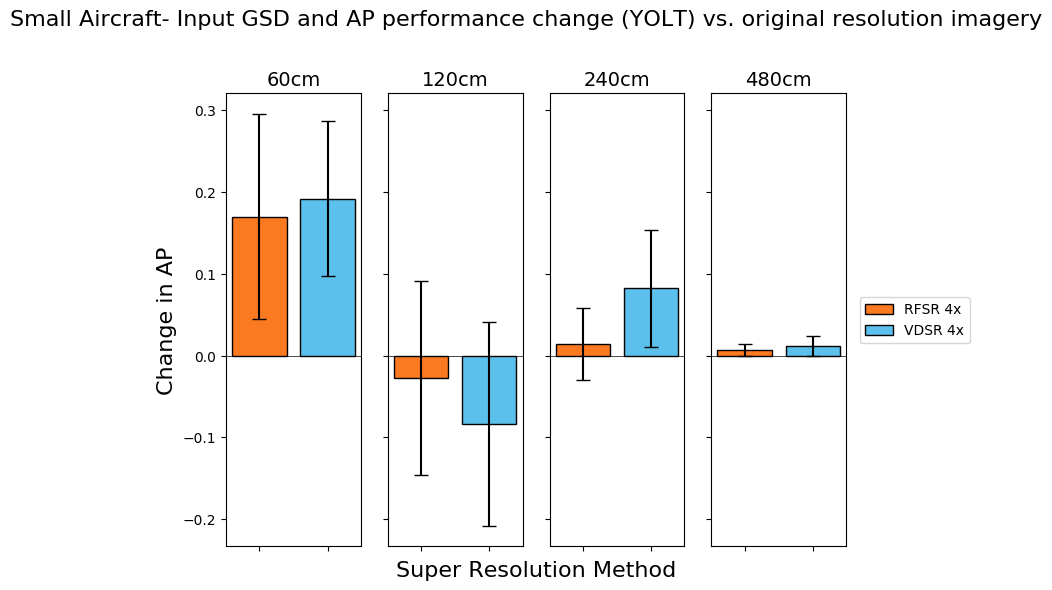}
\end{center}
\vspace{-4mm}%Put here to reduce too much white space after your table 
\caption{Performance change over original resolution (Figure \ref{fig:sair} using YOLT and 4x super-resolved data (Small Aircraft)).}
\end{figure}

\begin{table*}[h!]
%\resizebox{\textwidth}{!}{%
\scriptsize%\footnotesize%\small%\tiny
\begin{center}
\begin{tabular}{lllllllll}
\hline
Model & Data           & 30 cm                          & 60 cm                          & 120 cm                         & 240 cm                         & 480 cm                         \\ \hline\hline
YOLT  & Native         & $0.59 \pm 0.13$                & $0.55 \pm 0.11$                & $0.55 \pm 0.11$                & $0.11 \pm 0.05$                & $0.0 \pm 0.00$                \\
YOLT  & RFSR $2\times$ & $0.70 \pm 0.10$ (+0.7$\sigma$) & $0.64 \pm 0.12$ (+0.5$\sigma$) & $0.46 \pm 0.12$ (-0.6$\sigma$) & $0.18 \pm 0.06$ (+0.8$\sigma$) & $0.01 \pm 0.01$ (+1.1$\sigma$) \\
YOLT  & VDSR $2\times$ & $0.66 \pm 0.14$ (+0.4$\sigma$) & $0.52 \pm 0.10$ (-0.2$\sigma$) & $0.54 \pm 0.13$ (-0.1$\sigma$) & $0.11 \pm 0.04$ (+0.0$\sigma$) & $0.0 \pm 0.00$ (+0.3$\sigma$) \\
YOLT  & RFSR $4\times$ &                                & $0.72 \pm 0.13$ (+1.0$\sigma$) & $0.53 \pm 0.12$ (-0.2$\sigma$) & $0.13\pm0.04$ (+0.2$\sigma$)   & $0.01\pm0.01$ (+1.0$\sigma$)   \\
YOLT  & VDSR $4\times$ &                                & $0.75 \pm 0.09$ (+1.3$\sigma$) & $0.47 \pm 0.12$ (-0.5$\sigma$) & $0.20\pm0.07$ (+0.9$\sigma$)   & $0.01\pm0.01$ (+1.0$\sigma$)   \\ 
\hline
SSD   & Native         & 0.14 $\pm$ 0.04                & 0.14 $\pm$ 0.04                & 0.05 $\pm$ 0.03                & 0 $\pm$ 0                      & 0 $\pm$ 0                      \\
SSD   & RFSR $2\times$ & 0.05 $\pm$ 0.03 (-1.8$\sigma$) & 0.24 $\pm$ 0.07 (+1.2$\sigma$) & 0.04 $\pm$ 0.02 (-0.1$\sigma$) & 0 $\pm$ 0 (0$\sigma$)          & 0 $\pm$ 0 (0$\sigma$)          \\
SSD   & VDSR $2\times$ & 0.37 $\pm$ 0.12 (+1.8$\sigma$) & 0.15 $\pm$ 0.07 (+0.1$\sigma$) & 0.08 $\pm$ 0.03 (-1.2$\sigma$) & 0.01 $\pm$ 0.01 (+0.9$\sigma$) & 0 $\pm$ 0 (0$\sigma$) \\
\hline
\end{tabular}%}
\end{center}
\vspace{-4mm}%Put here to reduce too much white space after your table 
\caption{Performance for the small aircraft class  For RFSR and VDSR at each resolution we note the error and statistical difference from the baseline model (e.g. +0.5$\sigma$).}
\label{tab:det_perf}
\end{table*}

\newpage
\clearpage

\begin{figure}[ht]
\begin{center}
\includegraphics[width=0.7\linewidth]{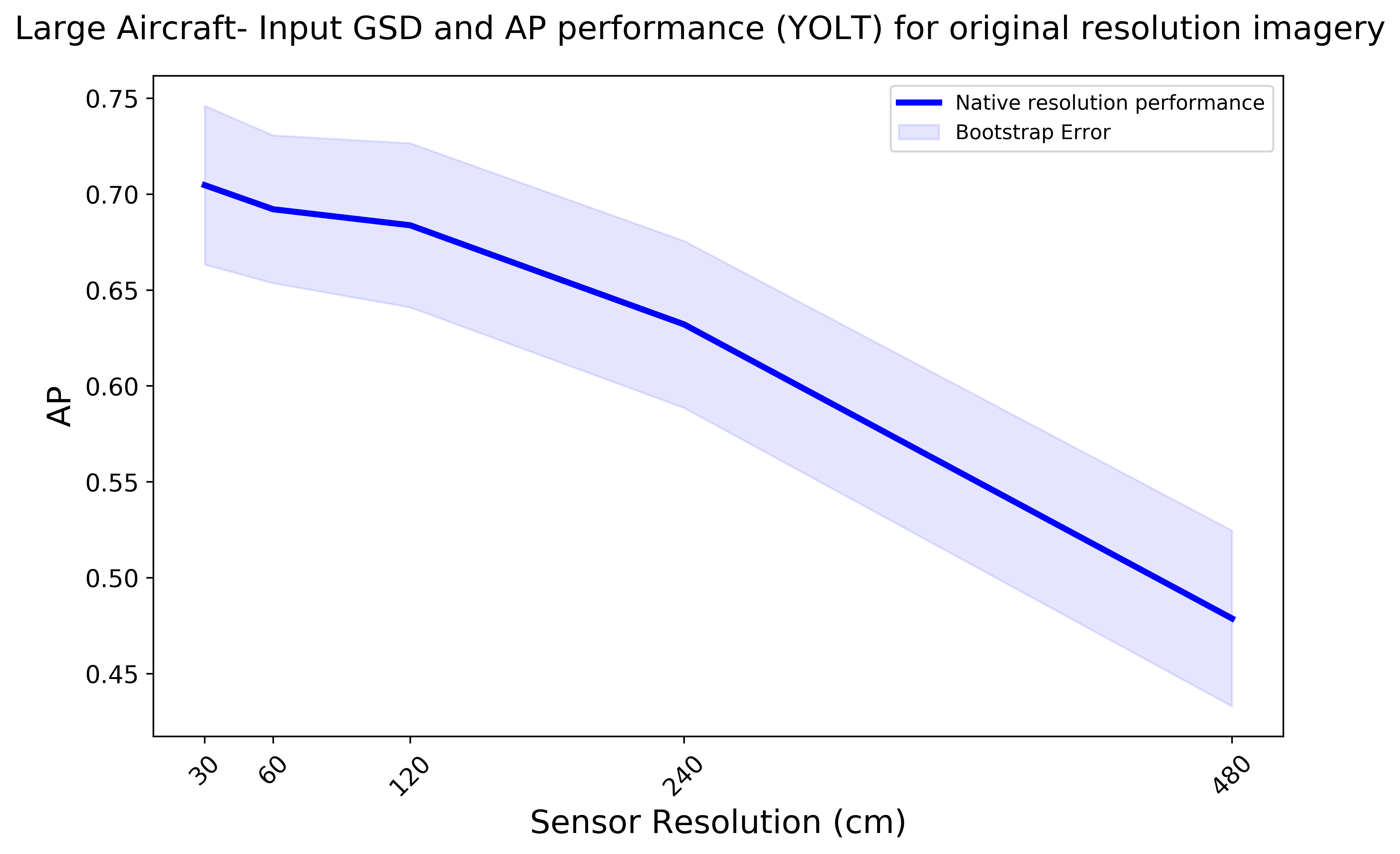}
\end{center}
\vspace{-4mm}%Put here to reduce too much white space after your table 
\caption{Performance of YOLT models on large aircraft as a function of sensor resolution.  The lower axis indicates the sensor resolution, with average precision plotted on the y-axis.}
\label{fig:lair}
\end{figure}

\begin{figure}[h!]
\begin{center}
\includegraphics[width=0.7\linewidth]{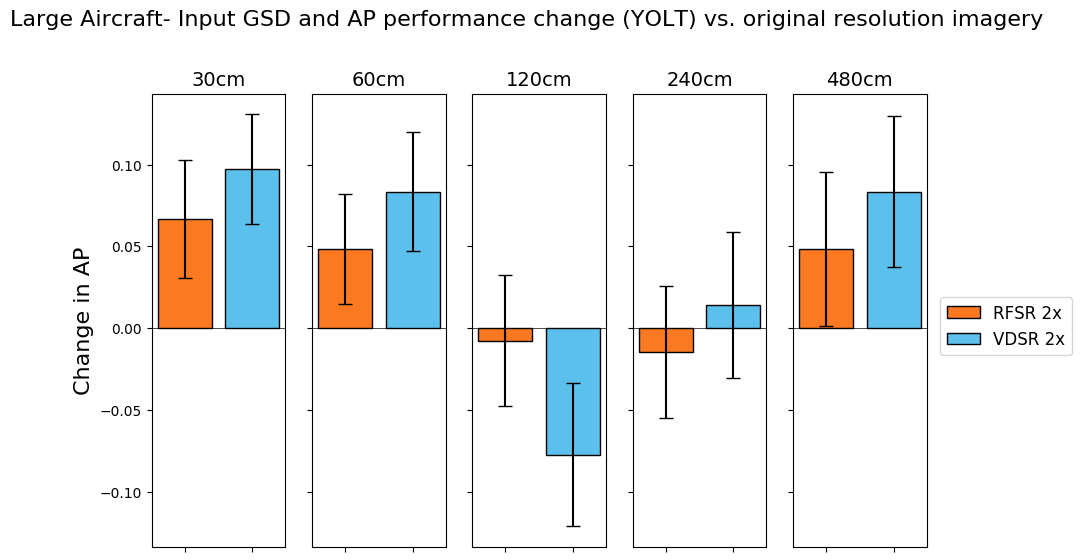}
\end{center}
\vspace{-4mm}%Put here to reduce too much white space after your table 
\caption{Performance change over original resolution (Figure \ref{fig:lair} using YOLT and 2x super-resolved data (Large Aircraft)).}
\end{figure}

\begin{figure}[h!]
\begin{center}
\includegraphics[width=0.7\linewidth]{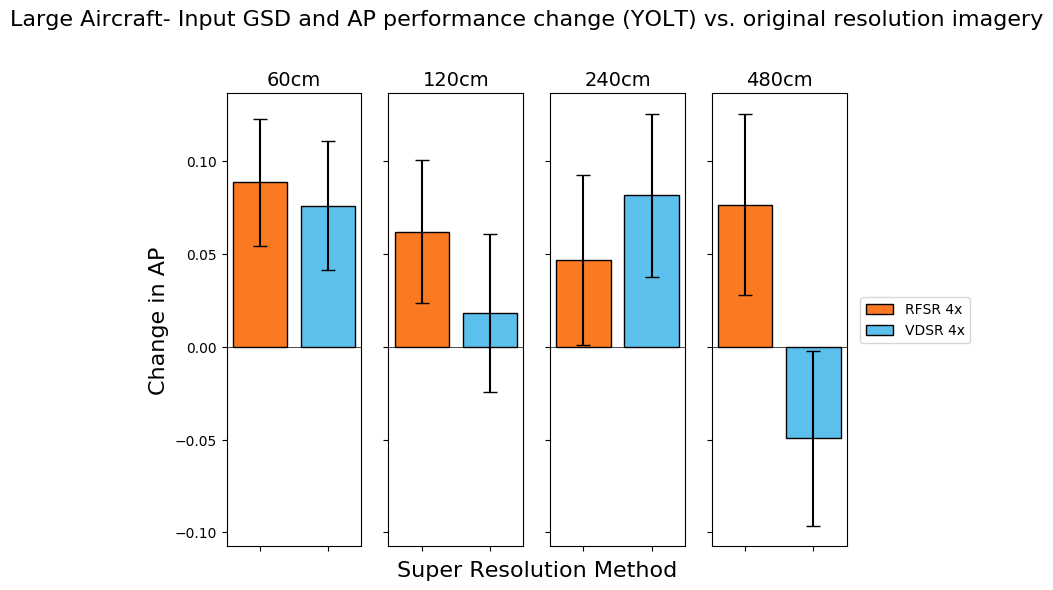}
\end{center}
\vspace{-4mm}%Put here to reduce too much white space after your table 
\caption{Performance change over original resolution (Figure \ref{fig:lair} using YOLT and 4x super-resolved data (Large Aircraft)).}
\end{figure}

\begin{table*}[hb]
%\resizebox{\textwidth}{!}{%
\scriptsize%\footnotesize%\small%\tiny
\begin{center}
\begin{tabular}{lllllllll}
\hline
Model & Data           & 30 cm                          & 60 cm                          & 120 cm                         & 240 cm                         & 480 cm                         \\ \hline\hline
YOLT  & Native         & $0.70 \pm 0.04$                & $0.69 \pm 0.04$                & $0.68 \pm 0.04$                & $0.63 \pm 0.04$                & $0.48 \pm 0.05$                \\
YOLT  & RFSR $2\times$ & $0.77 \pm 0.04$ (+1.2$\sigma$) & $0.74 \pm 0.03$ (+0.9$\sigma$) & $0.68 \pm 0.04$ (-0.1$\sigma$) & $0.62 \pm 0.04$ (-0.2$\sigma$) & $0.53 \pm 0.05$ (+0.7$\sigma$) \\
YOLT  & VDSR $2\times$ & $0.80 \pm 0.03$ (+1.8$\sigma$) & $0.78 \pm 0.04$ (+1.6$\sigma$) & $0.61 \pm 0.04$ (-1.3$\sigma$) & $0.65 \pm 0.04$ (+0.2$\sigma$) & $0.56 \pm 0.05$ (+1.3$\sigma$) \\
YOLT  & RFSR $4\times$ &                                & $0.78 \pm 0.03$ (+1.7$\sigma$) & $0.75 \pm 0.04$ (+1.1$\sigma$) & $0.68\pm0.05$ (+0.7$\sigma$)   & $0.56\pm0.05$ (+1.1$\sigma$)   \\
YOLT  & VDSR $4\times$ &                                & $0.77 \pm 0.03$ (+1.5$\sigma$) & $0.70 \pm 0.04$ (+0.3$\sigma$) & $0.71\pm0.04$ (+1.3$\sigma$)   & $0.43\pm0.05$ (-0.7$\sigma$)   \\ 
\hline
SSD   & Native         & 0.7 $\pm$ 0.04                 & 0.68 $\pm$ 0.04                & 0.69 $\pm$ 0.04                & 0.36 $\pm$ 0.05                & 0 $\pm$ 0                      \\
SSD   & RFSR $2\times$ & 0.58 $\pm$ 0.03 (-2.4$\sigma$) & 0.67 $\pm$ 0.04 (-0.2$\sigma$) & 0.65 $\pm$ 0.04 (-0.7$\sigma$) & 0.56 $\pm$ 0.05 (+2.7$\sigma$) & 0.34 $\pm$ 0.05 (+6.9$\sigma$) \\
SSD   & VDSR $2\times$ & 0.53 $\pm$ 0.05 (-2.7$\sigma$) & 0.63 $\pm$ 0.04 (-1$\sigma$)   & 0.65 $\pm$ 0.04 (-0.5$\sigma$) & 0.55 $\pm$ 0.04 (+2.8$\sigma$) & 0.41 $\pm$ 0.05 (+8.6$\sigma$) \\
\hline
\end{tabular}%}
\end{center}
\vspace{-4mm}%Put here to reduce too much white space after your table 
\caption{Performance for the large aircraft class.  For RFSR and VDSR at each resolution we note the error and statistical difference from the baseline model (e.g. +0.5$\sigma$). }
\label{tab:det_perf}
\end{table*}

\newpage
\clearpage

\begin{figure}[ht]
\begin{center}
\includegraphics[width=0.7\linewidth]{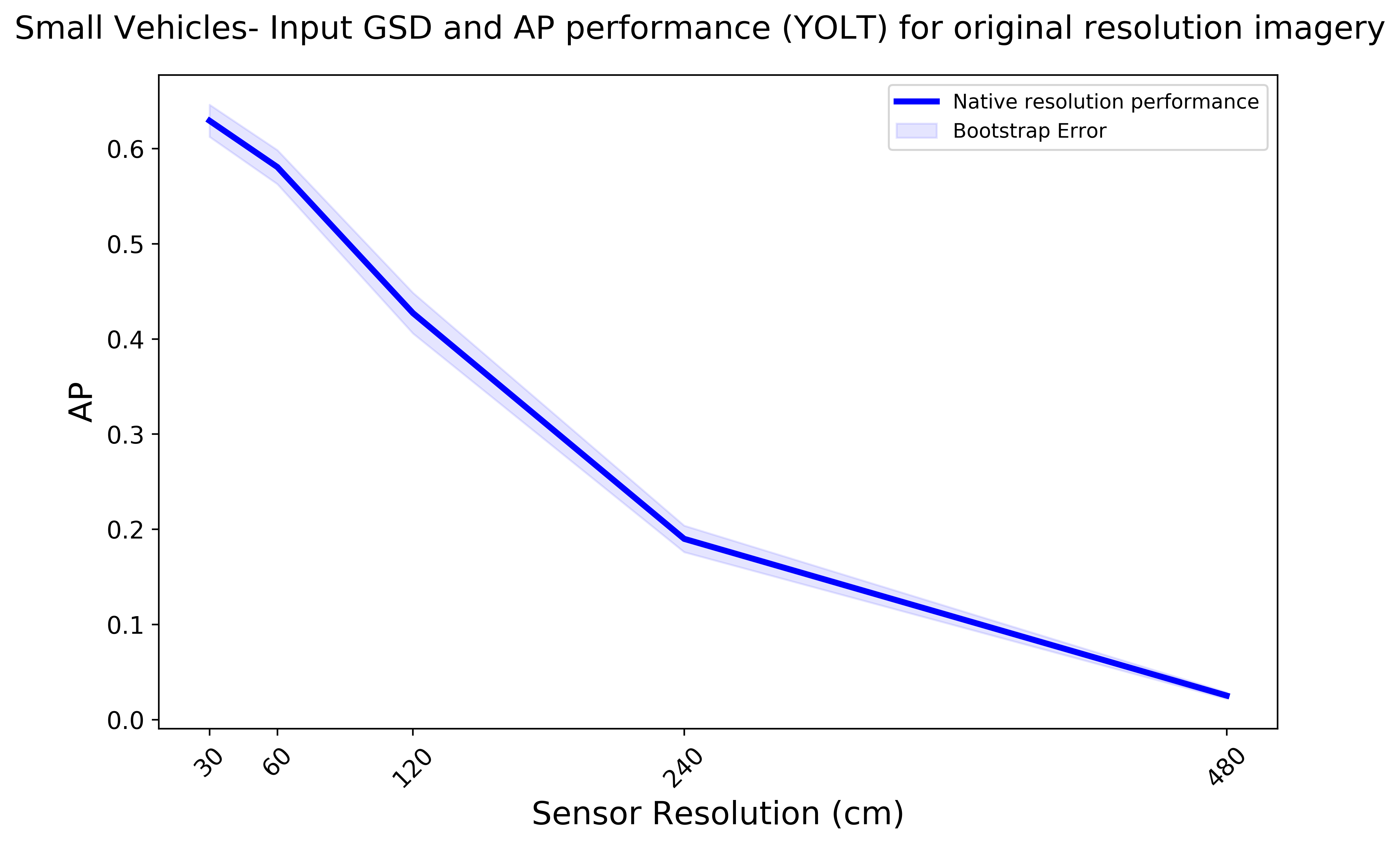}
\end{center}
\vspace{-4mm}%Put here to reduce too much white space after your table 
\caption{Performance of YOLT models on small vehicles as a function of sensor resolution.  The lower axis indicates the sensor resolution, with average precision plotted on the y-axis.}
\label{fig:sv}
\end{figure}

\begin{figure}[h!]
\begin{center}
\includegraphics[width=0.7\linewidth]{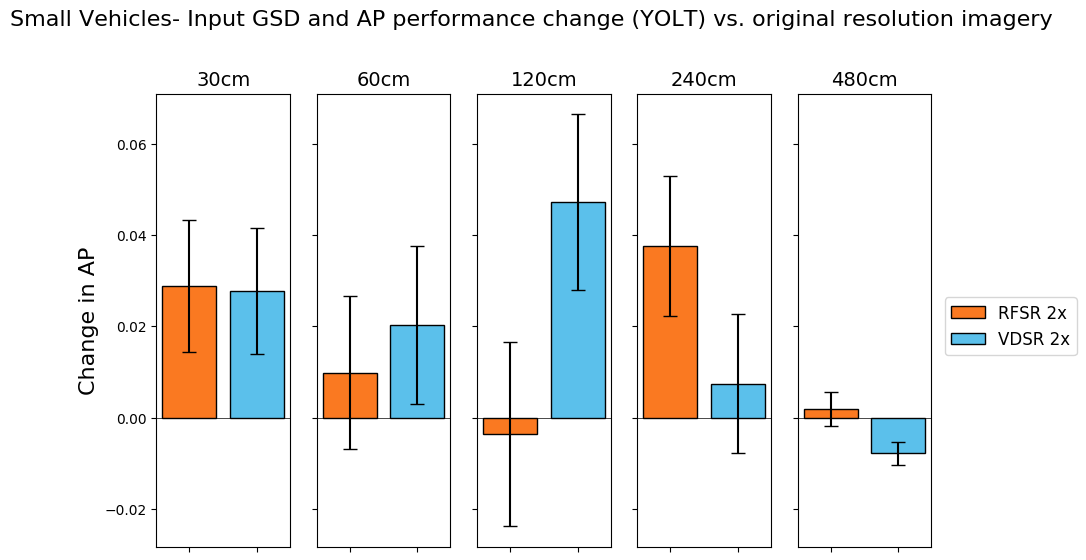}
\end{center}
\vspace{-4mm}%Put here to reduce too much white space after your table 
\caption{Performance change over original resolution (Figure \ref{fig:sv} using YOLT and 2x super-resolved data (Small Vehicles)).}
\end{figure}

\begin{figure}[h!]
\begin{center}
\includegraphics[width=0.7\linewidth]{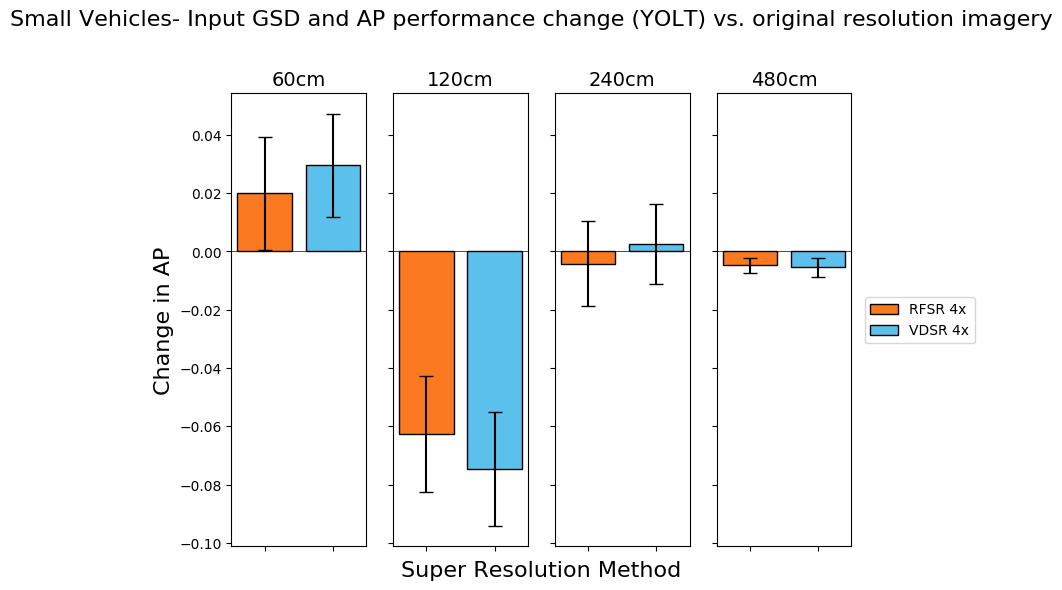}
\end{center}
\vspace{-4mm}%Put here to reduce too much white space after your table 
\caption{Performance change over original resolution (Figure \ref{fig:sv} using YOLT and 4x super-resolved data (Small Vehicles)).}
\end{figure}

\begin{table*}[hb]
%\resizebox{\textwidth}{!}{%
\scriptsize%\footnotesize%\small%\tiny
\begin{center}
\begin{tabular}{lllllllll}
\hline
Model & Data           & 30 cm                          & 60 cm                          & 120 cm                         & 240 cm                         & 480 cm                         \\ \hline\hline
YOLT  & Native         & $0.63 \pm 0.02$                & $0.58 \pm 0.02$                & $0.43 \pm 0.02$                & $0.19 \pm 0.01$                & $0.03 \pm 0.00$                \\
YOLT  & RFSR $2\times$ & $0.66 \pm 0.01$ (+1.3$\sigma$) & $0.59 \pm 0.02$ (+0.4$\sigma$) & $0.42 \pm 0.02$ (-0.1$\sigma$) & $0.23 \pm 0.02$ (+1.8$\sigma$) & $0.03 \pm 0.00$ (+0.4$\sigma$) \\
YOLT  & VDSR $2\times$ & $0.66 \pm 0.01$ (+1.7$\sigma$) & $0.60 \pm 0.02$ (+0.8$\sigma$) & $0.47 \pm 0.02$ (+1.6$\sigma$) & $0.20 \pm 0.02$ (+0.4$\sigma$) & $0.02 \pm 0.00$ (-1.8$\sigma$) \\
YOLT  & RFSR $4\times$ &                                & $0.56 \pm 0.03$ (+1.9$\sigma$) & $0.40 \pm 0.03$ (-0.2$\sigma$) & $0.23\pm0.01$ (+0.8$\sigma$)   & $0.12\pm0.01$ (+1.4$\sigma$)   \\
YOLT  & VDSR $4\times$ &                                & $0.59 \pm 0.02$ (+3.0$\sigma$) & $0.39 \pm 0.03$ (-0.6$\sigma$) & $0.25\pm0.02$ (+1.7$\sigma$)   & $0.10\pm0.01$ (-0.3$\sigma$)   \\ 
\hline
SSD   & Native         & 0.48 $\pm$ 0.01                 & 0.46 $\pm$ 0.02                & 0.27 $\pm$ 0.02                & 0.04 $\pm$ 0.01                & 0 $\pm$ 0                   \\
SSD   & RFSR $2\times$ & 0.75 $\pm$ 0.01 (+14.3$\sigma$) & 0.49 $\pm$ 0.02 (+1.2$\sigma$) & 0.31 $\pm$ 0.02 (+1.7$\sigma$) & 0.07 $\pm$ 0.01 (+2.4$\sigma$) & 0 $\pm$ 0 (-2.5$\sigma$)    \\
SSD   & VDSR $2\times$ & 0.72 $\pm$ 0.02 (+11.1$\sigma$) & 0.49 $\pm$ 0.02 (+1.2$\sigma$) & 0.38 $\pm$ 0.02 (-3.2$\sigma$) & 0.08 $\pm$ 0.01 (+3.4$\sigma$) & 0 $\pm$ 0 (-2.5$\sigma$)  \\ 
\hline
\end{tabular}%}
\end{center}
%\vspace{-4mm}%Put here to reduce too much white space after your table 
\caption{Performance for the small vehicle class.  For RFSR and VDSR at each resolution we note the error and statistical difference from the baseline model (e.g. +0.5$\sigma$).}
\label{tab:det_perf}
\end{table*}

\newpage
\clearpage

\begin{figure}[ht]
\begin{center}
\includegraphics[width=0.7\linewidth]{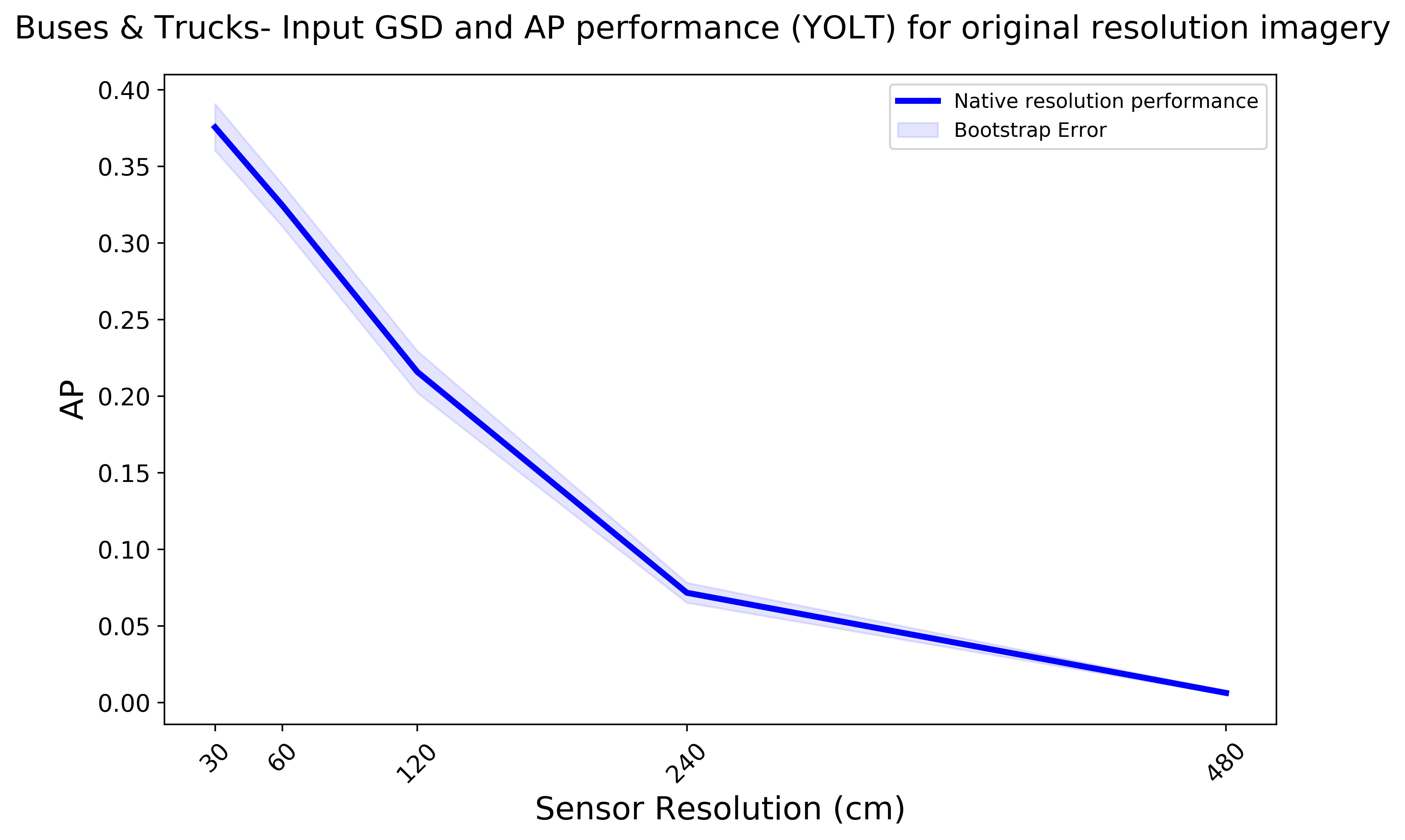}
\end{center}
\vspace{-4mm}%Put here to reduce too much white space after your table 
\caption{Performance of YOLT models on buses and trucks as a function of sensor resolution.  The lower axis indicates the sensor resolution, with average precision plotted on the y-axis.}
\label{fig:BT}
\end{figure}

\begin{figure}[h!]
\begin{center}
\includegraphics[width=0.7\linewidth]{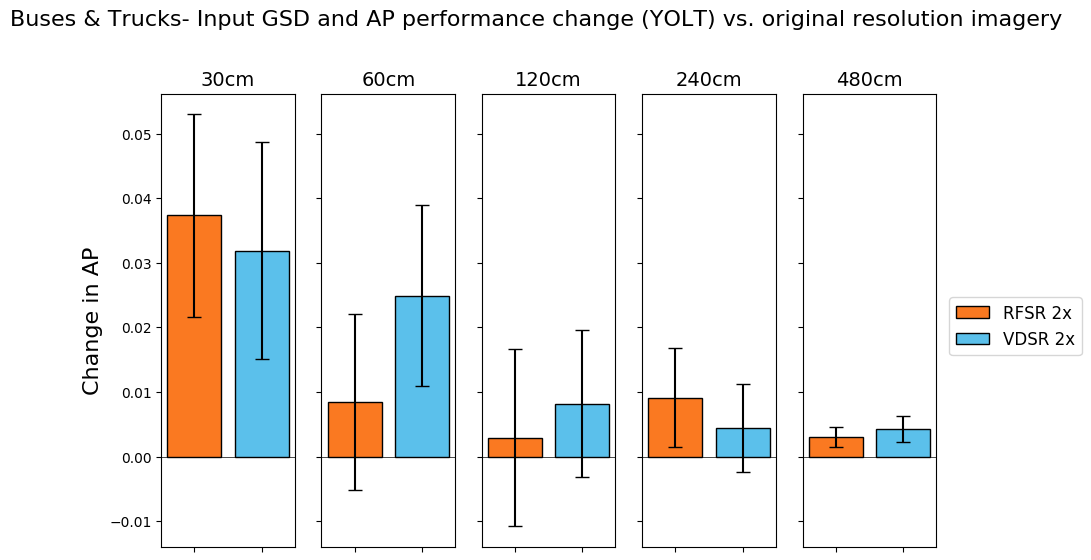}
\end{center}
\vspace{-4mm}%Put here to reduce too much white space after your table 
\caption{Performance change over original resolution (Figure \ref{fig:BT} using YOLT and 2x super-resolved data (Buses and Trucks)).}
\end{figure}

\begin{figure}[h!]
\begin{center}
\includegraphics[width=0.7\linewidth]{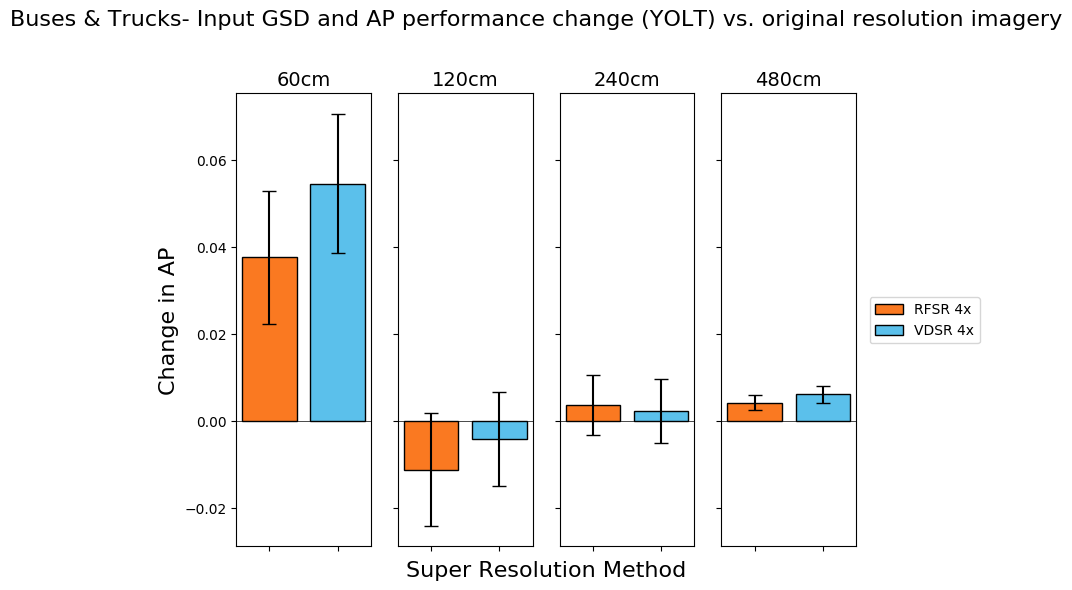}
\end{center}
\vspace{-4mm}%Put here to reduce too much white space after your table 
\caption{Performance change over original resolution (Figure \ref{fig:BT} using YOLT and 4x super-resolved data (Buses and Trucks)).}
\end{figure}

\begin{table*}[hb]
%\resizebox{\textwidth}{!}{%
\scriptsize%\footnotesize%\small%\tiny
\begin{center}
\begin{tabular}{lllllllll}
\hline
Model & Data           & 30 cm                          & 60 cm                          & 120 cm                         & 240 cm                         & 480 cm                         \\ \hline\hline
YOLT  & Native         & $0.38 \pm 0.02$                & $0.32 \pm 0.01$                & $0.22 \pm 0.01$                & $0.07 \pm 0.01$                & $0.01 \pm 0.00$                \\
YOLT  & RFSR $2\times$ & $0.41 \pm 0.02$ (+1.7$\sigma$) & $0.33 \pm 0.01$ (+0.4$\sigma$) & $0.22 \pm 0.03$ (+0.2$\sigma$) & $0.08 \pm 0.01$ (+0.9$\sigma$) & $0.01 \pm 0.00$ (+1.5$\sigma$) \\
YOLT  & VDSR $2\times$ & $0.41 \pm 0.02$ (+1.4$\sigma$) & $0.35 \pm 0.01$ (+1.3$\sigma$) & $0.22 \pm 0.01$ (+0.5$\sigma$) & $0.08 \pm 0.01$ (+0.5$\sigma$) & $0.01 \pm 0.00$ (+1.8$\sigma$) \\
YOLT  & RFSR $4\times$ &                                & $0.36 \pm 0.02$ (+1.8$\sigma$) & $0.20 \pm 0.01$ (-0.6$\sigma$) & $0.08\pm0.01$ (+0.4$\sigma$)   & $0.01\pm0.00$ (+2.0$\sigma$)   \\
YOLT  & VDSR $4\times$ &                                & $0.38 \pm 0.02$ (+2.6$\sigma$) & $0.21 \pm 0.01$ (-0.2$\sigma$) & $0.07\pm0.01$ (+0.2$\sigma$)   & $0.01\pm0.00$ (+2.6$\sigma$)   \\ 
\hline
SSD   & Native         & 0.28 $\pm$ 0.01                & 0.16 $\pm$ 0.01                & 0.06 $\pm$ 0.01                & 0 $\pm$ 0                      & 0 $\pm$ 0                   \\
SSD   & RFSR $2\times$ & 0.26 $\pm$ 0.02 (-0.8$\sigma$) & 0.17 $\pm$ 0.01 (+0.9$\sigma$) & 0.14 $\pm$ 0.01 (+6.1$\sigma$) & 0 $\pm$ 0 (+3.2$\sigma$)       & 0 $\pm$ 0 (+0$\sigma$)      \\
SSD   & VDSR $2\times$ & 0.33 $\pm$ 0.02 (+2$\sigma$)   & 0.2 $\pm$ 0.01 (+2.7$\sigma$)  & 0.15 $\pm$ 0.01 (-0.5$\sigma$) & 0.01 $\pm$ 0 (+4.4$\sigma$)    & 0 $\pm$ 0 (+0$\sigma$)     \\
\hline
\end{tabular}%}
\end{center}
%\vspace{-4mm}%Put here to reduce too much white space after your table 
\caption{Performance for the truck and bus class.  For RFSR and VDSR at each resolution we note the error and statistical difference from the baseline model (e.g. +0.5$\sigma$).}
\label{tab:det_perf}
\end{table*}

\end{document}